\documentclass{article}
\usepackage{arxiv}
\usepackage[utf8]{inputenc} 
\usepackage[T1]{fontenc}    
\usepackage{hyperref}       
\usepackage{url}           
\usepackage{booktabs}       
\usepackage{amsfonts}   
\usepackage{nicefrac}       
\usepackage{microtype}      
\usepackage{lipsum}
\usepackage{graphicx}
\usepackage{subfig}
\usepackage{amsmath}
\usepackage{multirow}
\usepackage{authblk}
\usepackage{enumitem}
\usepackage{algorithm}
\usepackage{algorithmic}  
\graphicspath{ {./images/} }
\newtheorem{definition}{Definition}

\title{A Review of Designs and Applications of Echo State Networks}

\author[1,2]{Chenxi Sun}
\author[1,2]{Moxian Song}
\author[3,4]{Shenda Hong}
\author[1,2]{Hongyan Li \thanks{Corresponding author: leehy@pku.edu.cn}}
\affil[1]{Key Laboratory of Machine Perception (Ministry of Education), Peking University, Beijing, China.}
\affil[2]{School of Electronics Engineering and Computer Science, Peking University, Beijing, China.}
\affil[3]{National Institute of Health Data Science, Peking University, Beijing, China.}
\affil[4]{Institute of Medical Technology, Health Science Center, Peking University, Beijing, China.}

\begin{document}
\maketitle
\begin{abstract}

Recurrent Neural Networks (RNNs) have demonstrated their outstanding ability in sequence tasks and have achieved state-of-the-art in wide range of applications, such as industrial, medical, economic and linguistic. Echo State Network (ESN) is simple type of RNNs and has emerged in the last decade as an alternative to gradient descent training based RNNs. ESN, with a strong theoretical ground, is practical, conceptually simple, easy to implement. It avoids non-converging and computationally expensive in the gradient descent methods. Since ESN was put forward in 2002, abundant existing works have promoted the progress of ESN, and the recently introduced Deep ESN model opened the way to uniting the merits of deep learning and ESNs. Besides, the combinations of ESNs with other machine learning models have also overperformed baselines in some applications. However, the apparent simplicity of ESNs can sometimes be deceptive and successfully applying ESNs needs some experience. Thus, in this paper, we categorize the ESN-based methods to basic ESNs, DeepESNs and combinations, then analyze them from the perspective of theoretical studies, network designs and specific applications. Finally, we discuss the challenges and opportunities of ESNs by summarizing the open questions and proposing possible future works.

\end{abstract}

\section{Introduction} \label{sec:Introduction}

In the last decades, Recurrent Neural Networks (RNNs) based methods demonstrated their outstanding ability in time-series prediction tasks and have become very attractive for their potentially wide range of applications, such as biological, medical, linguistic, social and economic. RNNs represent a very powerful generic tool, integrating both large dynamical memory and highly adaptable computational capabilities. They are the Machine Learning (ML) model most closely resembling biological brains, the substrate of natural intelligence. Meanwhile, deep RNN are able to develop in their internal states a multiple time-scales representation of the temporal information, a much desired feature. 

In the context of deep learning, error backpropagation (BP) \cite{BPerror} is to this date one of the most important achievements in RNNs training, but only with a partial success. One of the limitations is that bifurcations can make training non-converging, computationally expensive and poor local minima \cite{BPerror2}. Moreover, the exploding gradient problem often occurs in the training process, where the stable prediction performance cannot be easily ensured \cite{DBLP:journals/pr/ChatzisD12}. Some developments in BP for RNNs \cite{DBLP:conf/icml/MartensS11} perform better on problems which require long memory, but it's hard for BP RNN training, unless networks are specifically designed to deal with them \cite{DBLP:series/lncs/Lukosevicius12}. Long short-memory (LSTM) and gated recurrent unit (GRU) are advanced designs to mitigate shortcomings of RNN. But when the length of input sequence exceeds a certain limit, the gradient will still disappear. Meanwhile, each LSTM cell have four full connection layers, if the time span of LSTM is large and the network is very deep, the calculation will be very heavy and time-consuming. Further, too many parameters will lead to over fitting risk.

Reservoir computing (RC) has emerged in the last decade as an alternative to gradient descent methods for training RNNs with a strong theoretical ground \cite{DBLP:conf/nips/Jaeger02,DBLP:series/sci/TinoHB07,2011Architectural,DBLP:journals/corr/abs-1712-04323,DBLP:journals/nn/YildizJK12}. Echo State Network (ESN) \cite{DBLP:conf/nips/Jaeger02} is one of the key RC. ESNs are practical, conceptually simple, and easy to implement. ESNs employ the multiple high-dimensional projection in the large number of states of the reservoir with strong nonlinear mapping capabilities, to capture the dynamics of the input. This basic idea was first clearly spelled out in a neuroscientific model of the corticostriatal processing loop \cite{DBLP:journals/pr/ChatzisD12}. ESNs enjoy, under mild conditions, the so-called echo state property \cite{DBLP:conf/nips/Jaeger02}, that ensures that the effect of the initial condition vanishes after a finite transient. The inputs with more similar short-term history will evoke closer echo states, which ensure the dynamical stability of the reservoir. 

Recently, the introduction of the Deep Echo State Network (DeepESN) model \cite{DBLP:conf/esann/GallicchioM16,Gallicchio2017Deep} allowed to study the properties of layered RNN architectures separately from the learning aspects. Remarkably, such studies pointed out that the structured state space organization with multiple time-scales dynamics in deep RNNs is intrinsic to the nature of compositionality of recurrent neural modules. The interest in the study of the DeepESN model is hence twofold. On the one hand, it allows to shed light on the intrinsic properties of state dynamics of layered RNN architectures. On the other hand it enables the design of extremely efficiently trained deep neural networks for temporal data.

Meanwhile, with the development of deep learning, a variety of neural network structures have been proposed, like auto-encoder (AE) \cite{Vincent2008Extracting}, generative adversarial networks (GAN) \cite{DBLP:conf/nips/GoodfellowPMXWOCB14}, convolutional neural networks (CNNs) \cite{DBLP:journals/corr/SimonyanZ14a} and 
restricted Boltzmann machine (RBM) \cite{DBLP:journals/cogsci/AckleyHS85}. ESNs have combined these different structures and achieved state-of-the-art performance in specific tasks and different applications, such as industrial \cite{DBLP:conf/ijcnn/MansoorGM20}, medical \cite{DBLP:conf/cinc/AlfarasVG19}, financial \cite{DBLP:conf/iconip/LiuSLFCZD18} and robotics with reinforcement learning \cite{DBLP:conf/cifer/MacielGSB14,DBLP:conf/aaai/WhitmanBTC20}.

ESNs from their beginning proved to be a highly practical approach to RNN training. It is conceptually simple and computationally inexpensive. It reinvigorated interest in RNNs, by making them accessible to wider audiences. However, the apparent simplicity of ESNs can sometimes be deceptive. Successfully applying ESNs needs some experience. Thus, we summarize the advancements in the development, analysis and applications of ESNs to summarize the experience of existing work.

The rest of the paper is organized as follows. Section \ref{sec:ESNs} analyzes the basic ESNs from the basic definitions to specific designs. Section \ref{sec:DeepESNs} introduces the DeepESNs with its different structures. Section \ref{sec:xESNs} describes the existing combinations of ESN and other machine learning methods. Section \ref{sec:Applications} gives the benchmarks, the data-driven and the real-ward applications of ESNs. Section \ref{sec:Discussions} and \ref{sec:Conclusion} raise the challenges, opportunities and make conclusions. 

\section{Basic echo state networks} \label{sec:ESNs}

\subsection{Preliminaries} \label{sec:ESNs_Preliminaries}

\begin{figure*}[t]
\centerline{\includegraphics[width=0.6\linewidth]{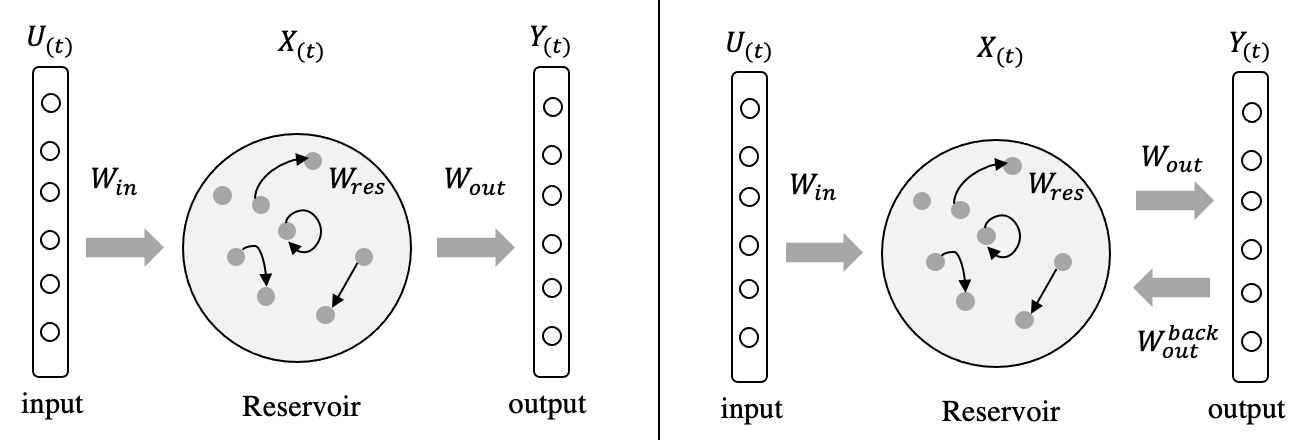}}
\caption{Basic echo state networks}
\label{fig:ESN}
\end{figure*}

\begin{definition}[Echo State Networks (ESNs)] \label{def:ESNs}
Echo state networks are a type of fast and efficient recurrent neural network.
A typical ESN consists of an input layer, a recurrent layer, called reservoir, with a large number of sparsely connected neurons, and an output layer. The connection weights of the input layer and the reservoir layer are fixed after initialization, and the output weights are trainable and obtained by solving a linear regression problem. 
\end{definition}

In an ESN, $u(t)\in R^{D\times 1}$ denotes the input value at time $t$ of the time series, $x(t)\in R^{N\times 1}$ denotes the state of the reservoir at time $t$. $y(t)\in R^{M\times 1}$ denotes the output value. $W_{in}\in R^{N\times D}$ represents the connection weights between the input layer and the hidden layer, $W_{res}\in R^{N\times N}$ notes the connection weights in hidden layer. The state transition equation is in Equation \ref{eq:ESN_trans}, where $f$ is a nonlinear function such as $\tan$ and sigmoid. 
\begin{equation} \label{eq:ESN_trans}
\begin{aligned}
x(t)&=f (W_{in}u(t)+W_{res}x(t-1))\\
y(t) &= W_{out}x(t)   
\end{aligned}
\end{equation}

Only parameters of readout weights $W_{out}$ are subject to training. Which can obtain a closed-form solution by extremely fast algorithms such as ridge regression. Assuming the internal states and desired outputs are stored in $X$ and $Y$. By training to find a solution to the least squares problem in Equation \ref{eq:ESN_objective}, the calculation formula of the readout weights $W_{out}$ is got in Equation \ref{eq:ESN_solve}.

\begin{equation} \label{eq:ESN_objective}
\min_{W_{out}}||W_{out}X-Y||^2_2
\end{equation}

\begin{equation} \label{eq:ESN_solve}
W_{out} = Y\cdot X^{-1}
\end{equation}

Before training phase, three are three main hyper-parameters of ESNs need to initialize - $w^{in}$, $\alpha$ and $\rho(W_{res})$. 

\begin{itemize} [leftmargin=10 pt]
    \item $w^{in}$ is an input-scaling parameter, the elements in $W_{in}$ are commonly randomly initialized from a uniform distribution in $[-w^{in},w^{in}]$. 
    \item $\alpha$ is the sparsity parameter of $W_{res}$, denoting the proportion of non-zero elements in matrix. 
    \item $\rho(W_{res})$ is the spectral radius parameter (the largest eigenvalue in absolute value) of $W_{res}$. $W_{res}$ initialize from a matrix $W$, where the elements of $W$ are generated randomly in $[-1,1]$ and $\lambda_{max}(W)$ is the largest eigenvalue.
\begin{equation} \label{eq:spectral_radius}
W_{res} =\rho(W_{res}) \cdot \frac{W}{\lambda_{max}(W)}
\end{equation}
\end{itemize}

The extreme training efficiency of the ESN approach derives from the fact that only the readout weights are trained, while the weights of input and reservoir are initialized under certain conditions and then are left untrained. 

After training phase, ESNs can give predictions $\hat{Y}$ of new data $\hat{U}$ with the trained $W_{out}$ by Equation \ref{eq:ESN_pre}.

\begin{equation} \label{eq:ESN_pre}
\begin{aligned}
\hat{X}_{t}&=f(W_{in}\hat{U}_{t}+W_{res}\hat{X}_{t-1})\\
\hat{Y}_{t}&= W_{out}\hat{X}_{t}
\end{aligned}
\end{equation}

Assuming that the size of the reservoir are fixed by N, The input time series data has T-length D-dimension. For the typical reservoir computing Equation \ref{eq:ESN_trans}, its complexity is computed in Equation \ref{eq:ESN_complexity}.

\begin{equation} \label{eq:ESN_complexity}
C_{typical}=\mathcal{O}(\alpha TN^{2}+TND)
\end{equation}

A pivotal property of ESNs is Echo State Property (ESP), which characterizes valid ESN dynamics. ESP essentially describes that the states of reservoir should asymptotically depend only on the driving input signal, which means the state is an echo of the input, meanwhile, and the influence of initial conditions should progressively vanish with time. Briefly, according to Jeager's paper \cite{DBLP:conf/nips/Jaeger02}, the ESN is that every echo state vector $x$ is uniquely determined for every input sequence $u$. It implies that the nearby echo states have the similar input histories. 

\begin{definition}[Echo State Property (ESP)] \label{def:ESP} A network with state transition equation $F$ with has the echo state property if for each input sequence $U=[u(1),u(2),...u(N)]$ and all couples of initial states $x$, $x'$, it could hold the condition in Equation \ref{eq:ESP}. 
\end{definition}

\begin{equation} \label{eq:ESP}
||F(U,x)-F(U,x')||\rightarrow 0 \quad \textit{as}\quad N \rightarrow \infty
\end{equation}

The existence of ESP are verified by a necessary condition and a sufficient condition with spectral radius $\rho(W_{res})$ and the largest singular value $\overline{\sigma} (W_{res})=||W||_{2}$: 

\begin{equation} \label{eq:ESN_condition}
\begin{aligned}
&\textit{echo state property}\Rightarrow \rho(W_{res})<1\\
&\overline{\sigma}(W_{res})<1 \Rightarrow \textit{echo state property}
\end{aligned}
\end{equation}

Two conditions are commonly used in ESN literature for reservoir initialization \cite{DBLP:journals/csr/LukoseviciusJ09}. In recent years, the conditions for the ESP have been further investigated and refined in successive contributions \cite{ DBLP:journals/tnn/BuehnerY06, DBLP:journals/neco/ManjunathJ13, DBLP:journals/nn/YildizJK12, DBLP:journals/nn/WainribG16,DBLP:journals/ijon/YangQHW18}. For example, basd on the two basic conditions, Buehner et al. \cite{DBLP:journals/tnn/BuehnerY06} provided a rigorous bound for guaranteeing asymptotic stability of ESN, which was required in the applications where stability is required. Authors extended L2-norm $||W_{res}||_{2}$ to the D-norm $||W_{res}||_{D}=\overline{\sigma} (DW_{res}D^{-1})$. They connected two conditions of echo state property by $||W_{res}||_{D}=\rho(W_{res})+\epsilon$ and the sufficient condition changes to Equation \ref{eq:ESN_condition2}.

\begin{equation} \label{eq:ESN_condition2}
\overline{\sigma}D(W_{res}D^{-1})<1 \Rightarrow \textit{echo state property}
\end{equation}

Short-Term Memory (STM) denotes the memory effects connected with the transient activation dynamics of networks. Memory capacity (MC) is a quantitative measure of STM. MC is calculated by the determination coefficient of $W_{out}^{k}$, $d[W_{out}^{k}](u(n-k),y_{k}(n))=\frac{cov^{2}(u(n-k),y_{k}(n))}{\sigma^{2}(u(n))\sigma^{2}(y_{k}(n))}$ when training the ESN to generate at its output units delayed versions $u(n-k)$. The STM capacity of the network according to MC is described in Equation \ref{eq:MC}

\begin{equation} \label{eq:MC}
\begin{aligned}
&MC=\sum_{k=1}^{\infty}MC_{k}\\
&MC_{k}=\max_{W_{out}^{k}}d[W_{out}^{k}](u(n-k),y_{k}(n))
\end{aligned}
\end{equation}

\subsection{Designs} \label{sec:ESNs_Designs}

\subsubsection{Basic models} \label{sec:Basic_models}

In 2002, Jeager \cite{DBLP:conf/nips/Jaeger02} proposed and named the echo state networks (ESNs) at the first time. He introduced ESN as a computationally efficient and easy learning method based on artificial recurrent neural networks (RNNs). The schema of ESN is that we described in section \ref{sec:ESNs_Preliminaries}. The ESN were suitable for process chaotic time series, abounded nonlinear dynamical systems of the sciences and engineering. In 2004, Jeager et al. \cite{Jaeger2004Harnessing} applied ESN for predicting a chaotic time series and the accuracy were improved by a factor of 2400 over previous techniques.

In 2007, Jeager et al. \cite{2007Optimization} proposed another important structure, leaky-ESN, formalized in Equation \ref{eq:leaky-ESN}, where $a\in [0,1]$ is the reservoir neuron’s leaking rate and $\gamma$ is the compound gain. Leaky-ESN's reservoir used leaky integrator neurons and the leaking rate was determined and fixed after model selection. By the effect analysis and experiments in slow dynamic systems, noisy time series and time-warped dynamic patterns, leaky-ESN outperformed than ESN and could incorporate a time constant to act as a low-pass filter and the dynamics can be slowed down. 

\begin{equation} \label{eq:leaky-ESN}
x(t)=(1-a\gamma)x(t-1)+\gamma f(W_{in}u(t)+W_{res}x(t-1))
\end{equation}

ESNs with output feedbacks was also emerging. The hidden state is not only affected by the input of this time and the hidden state of the previous time, but also affected by the output of the previous time, shown as Figure \ref{fig:ESN} right. The state transition equation changed to Equation \ref{eq:back-ESN}.

\begin{equation} \label{eq:back-ESN}
x(t)=(1-a\gamma)x(t-1)+\gamma f(W_{in}u(t)+W_{res}x(t-1)+W^{back}_{out}y(t-1))
\end{equation}

Based on these original versions, many deformations have been proposed. 

\subsubsection{Designs of reservoir} \label{sec:Designs of reservoir}

\begin{figure*}[t]
\centerline{\includegraphics[width=\linewidth]{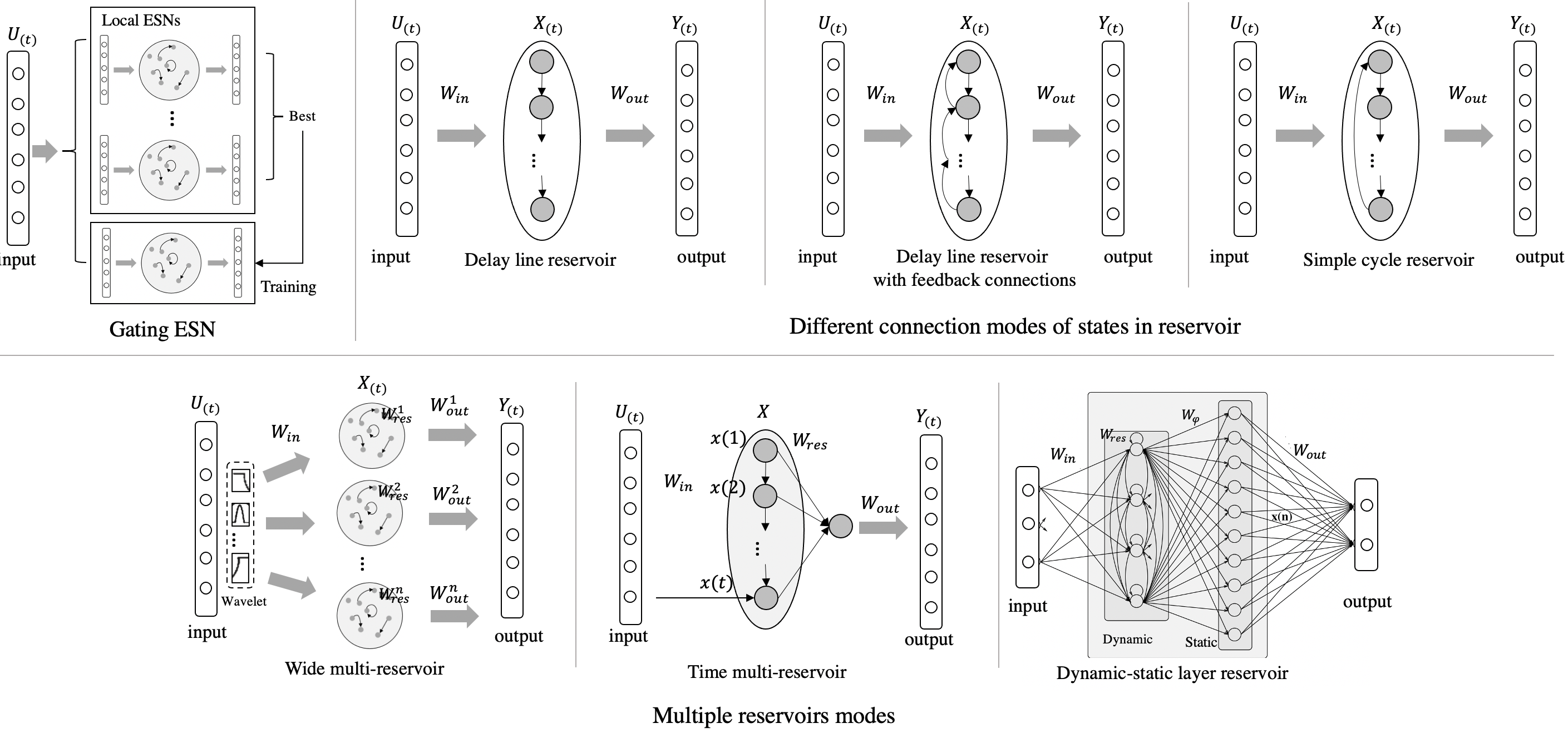}}
\caption{Designs of reservoir}
\label{fig:designs}
\end{figure*}

\begin{itemize}[leftmargin=10 pt]
    \item \textbf{Dynamic weights of reservoir}.
\end{itemize}

In the basic versions of ESNs, the weights in reservoir are fixed after initialization. However, the randomly initialized reservoir could not be always the most suitable one for the specific data and tasks, and it brings about complications with choosing starting parameters. Thus, the dynamic reservoir changing when processing data is a main focus when designing reservoir.

Mayer et al. \cite{DBLP:conf/bioadit/MayerB04} changed the random weights to adapted weights by a method of adapting the recurrent layer dynamics and self-prediction. The weights of reservoir changed to $\hat{W}_{res}$ by the output weight through output-prediction $W^{1}_{out}x_{t}=y(t)$ and self-prediction $W^{2}_{out}x_{t}=x_{t+1}$ in Equation \ref{eq:self-prediction-ESN}, where $\alpha$ is constant parameter. By adding self-prediction, the modified ESN could moderate the noise sensitivity.

\begin{equation} \label{eq:self-prediction-ESN}
\hat{W}_{res}=(1-\alpha)W_{res}+\alpha W^{2}_{out}
\end{equation}

Hajnal et al.\cite{DBLP:conf/icann/HajnalL06} introduced a notion of critical echo state networks (CESN), re-setiting $W_{res}$ to $\hat{W}_{res}$ by Equation \ref{eq:CESN}. In this case, $\hat{W}_{res}$ were able to approximate non-periodic dynamical systems, because the hidden layer were embedded by the input matrix $W_{in}$ and output matrix $W_{out}$, and they can modify finite cycles. 

\begin{equation} \label{eq:CESN}
\hat{W}_{res}=W_{res}+W_{in}W_{out}
\end{equation}

Fan et al. \cite{DBLP:conf/ijcnn/FanWJ17} proposed principle neuron reinforcement (PNR) algorithm to alter the strength of internal synapses within the reservoir by modifying the reservoir connections towards the goal of optimizing the neuronal dynamics. The PNR altered the connection according to the output weights of a pre-training stage, shown in Equation \ref{eq:PNR}. 

\begin{equation} \label{eq:PNR}
W_{res}(t)=(1+\rho (W_{res}))W_{res}(t-1), \quad \textit{if}\quad W_{out}>\xi
\end{equation}

Babinec et al. \cite{DBLP:conf/icann/BabinecP07} optimized ESN by the updating of $W_{res}$ with Anti-Oja’s learning (AO) by Equation \ref{eq:AO}, where $\eta$ is a small positive constant. AO changed the weights by decreasing correlation between neurons. It increased the diversity between hidden neurons and the internal state dynamics became much richer. 

\begin{equation} \label{eq:AO}
\begin{aligned}
&W_{res}(n+1)=W_{res}(n)-\Delta W_{res}(n)\\
&\Delta W_{res}(n)= \eta y(n)(x(n)-y(n)W_{res}(n))
\end{aligned}
\end{equation}

Boedecker et al. \cite{DBLP:conf/esann/BoedeckerOMA09} derived a new unsupervised learning rule based on intrinsic plasticity (IP) \cite{IP} to design weights of reservoir adapting dependency online. Then, Equation \ref{eq:ESN_trans} changed to Equation \ref{eq:IP}. Where $a$ is the gain vector and c is the bias vector. And the objective in Equation \ref{eq:ESN_objective} changed to Kullback-Leibler divergence $D_{KL}=\sum p(y)log\frac{p(y)}{\overline{p}(y)}$. Where $p(y)$ denotes the sampled output distribution of a reservoir neuron and $\overline{p}(y)$ is the desired output distribution with Laplace distribution $\overline{p}(y)=\frac{1}{2\sigma}e^{(-\frac{|x-\mu|}{\sigma})}$.

\begin{equation} \label{eq:IP}
y(t)=f(diag(a)W_{res}x(t-1)+diag(a)W_{in}u(t)+c)
\end{equation}

There are also many other designs about the dynamic connections. For example, Qiao et al. \cite{DBLP:journals/nca/QiaoWY19} proposed an adaptive lasso ESN (ALESN) to solve the problem of collinearity in randomly and sparsely connected reservoir. Wang et al. \cite{DBLP:journals/tnn/WangJH20} used local plasticity rules, that allow different neurons to use different parameters to learn local features, to replace the same type of plasticity rule used in the unmodified ESN, that makes the connections between neurons remain to be global. Babinec et al. \cite{DBLP:conf/iconip/BabinecP09} designed gating ESNs by combining some local expert ESNs with different spectral radius and used the them of local ESN with best prediction result to train gating ESN, shown in the first structure of Figure \ref{fig:designs}.

\begin{itemize}[leftmargin=10 pt]
    \item \textbf{Different connection modes of reservoir}.
\end{itemize}

There are three basic connections of states: 1) delay line reservoir (DLR), 2) delay line reservoir with feedback connections (DLRB) and 3) simple cycle reservoir (SCR) shown in Figure \ref{fig:designs}. In \cite{DBLP:journals/tnn/RodanT11}, authors tested the the minimal complexity of reservoir construction for obtaining competitive models and the memory capacity of such simplified reservoirs. They finally concluded that the memory capacity of SCR can be made arbitrarily close to the proved optimal value. 

Besides the basic connections, Sun et al. \cite{DBLP:journals/jzusc/SunCLCL12,DBLP:conf/cscwd/SunYZWX18} proposed the adjacent-feedback loop reservoir (ALR) based on SCR. ALR had a deterministic reservoir structure, in which the reservoir units were connected orderly in the loop manner and were also connected to the preceding one by adjacent feedback. In \cite{DBLP:journals/nn/CuiFCLL14}, authors also designed a circle reservoir structure with wavelet-neurons. To evaluate the different connections of reservoirs, 

For designing the size and topology of multiple reservoirs automatically, Growing ESN (GESN) \cite{DBLP:conf/ideal/Fan-junY16,DBLP:journals/tnn/QiaoLHL17} was proposed. It owned a growing reservoir with multiple sub-reservoirs by adding hidden units to the network group by group based on block matrix theory. And in \cite{DBLP:journals/asc/LiL19}, authors introduced using PSO and SVD algorithms to pre-train a growing ESN with multiple sub-reservoirs by optimizing singular values.

\begin{itemize}[leftmargin=10 pt]
    \item \textbf{Modes of multiple reservoirs}.
\end{itemize}

The above structures were built on the single reservoir, in practice, ESN can be extended with more reservoirs to achieve higher accuracy.

Chen et al. \cite{DBLP:conf/icmlc/ChenMP10} designed shared reservoir modular ESN (SRMESN) to first divided the neural state space into several modules of subspaces with independent output weight and then put the data belonging to each subspace into the same reservoir. Which means the ESNs own $n$ reservoirs rather than only one, shown in Figure \ref{fig:designs}. Many designs were proposed based on this. In \cite{DBLP:journals/tie/HanL13}, the inputs of this structure were utilized different wavelets of sequence; In \cite{DBLP:conf/itsc/SerLBV19}, authors built a stacking ESN by stacking ensemble of reservoir computing learners based that structure; In \cite{DBLP:journals/access/HanJLD19}, variational mode decomposition (VMD) was used to pick data as the input of different reservoirs. 

Meanwhile, another structure is shown in \ref{fig:designs}, the output $y(t)$ could also calculated from the multiple hidden states from time $1$ to $t$ \cite{DBLP:journals/tnn/XiaJHPM11}. However, the linear nature of the output layer may limit the capability of exploring the available information, since higher-order statistics of the signals are not taken into account. There are also other deigns of the output layer. In \cite{DBLP:conf/ijcnn/RachezH12}, authors used the nonlinear readout to represent the output probabilities; In \cite{DBLP:journals/nn/BoccatoLAZ12}, authors presented a Volterra filter structure and used principal component analysis to reduce the number of effective signals transmitted to the output layer.

Gallicchio and Micheli \cite{2011Architectural} pointed out that mapping the states of reservoir of ESN to the high-dimensional feature space was beneficial to both memory and good nonlinear mapping capabilities of the network, and proposed to extend the ESN with a random static nonlinear hidden layer. This type of structural variant is collectively referred to as $\varphi$-ESN, shown in Figure \ref{fig:designs}. In \cite{DBLP:journals/access/ZhangZT19}, An orthogonal least squares $\varphi$-ESN (OLS-$\varphi$-ESN) was proposed for generating the nodes of the extended static nonlinear hidden layer by applying incremental randomized learning method.

There are also other designs based on multiple reservoirs. In \cite{DBLP:journals/jfi/YaoW19}, authors proposed Broad-ESN through the unsupervised learning algorithm of restricted Boltzmann machine (RBM) to determine the number of reservoirs and aimed to fully reflect dynamic characteristics of a class of multivariate time series. Xue et al. \cite{DBLP:journals/nn/XueYH07} proposed decoupled ESN (DESN), used lateral inhibition unit to decouple the representation of state before input into multiple reservoirs. This structure have better robustness with respect to the random generation of reservoir weight matrix and feedback connections.

\begin{itemize}[leftmargin=10 pt]
    \item \textbf{Analysis of short-term memory in reservoirs}.
\end{itemize}

One of the most important properties of ESN is short-term memory, first introduced in \cite{JaegerShortTerm}. And all of the basic ESN including the above reservoir designs worked by modeling it. 

The topological structure of the reservoir relates to STM. Ma et al. \cite{DBLP:journals/asc/MaCWY14} first tested it. They concluded that the reservoir topology can be constructed according to the given memory span $r$, $r$ means the network can recover an input signal $r$ time steps ago. They also designed a model to convert $r$ to $W_{res}$. 

Different inputs also influence the STM. Barancok et al. \cite{DBLP:conf/icann/BarancokF14} calculated memory capacity (MC) of the networks for various input data, both random and structured, and showed that for uniformly distributed input data, the higher interval shift lead to higher MC; for structured data, depending on data properties. 

Meanwhile, the hyper-parameters $w^{in}$, $\rho(W_{res})$, $\alpha$ can also affect STM. Farkas et al. \cite{DBLP:journals/nn/FarkasBG16} systematically analyzed the MC from the perspective of that three hyper-parameters and their relations. They concluded that the optimal $w^{in}$ value moves with the reservoir size $N$, while $\rho(W_{res})$ and $\alpha$ approach for large $N$. And the reservoir size increases the MC sublinearly. 

Besides, Koryakin et al. \cite{DBLP:journals/nn/KoryakinLB12} proofed that the reservoir size and the output feedback strength crucially co-determines the likelihood of generating an effective ESN by experiments - somewhat smaller networks can yield better performance; the output feedback strength drives the dynamic reservoir but it can also block suitable reservoir dynamics. 

Further, Stockdill et al. \cite{DBLP:conf/ausai/StockdillN16} determine how leaking, loops, cycles and discrete time to influence the suitability of the reservoir. The results were that the single most important feature of a reservoir neural network is the discrete time step nature of input propagation; The loops and cycles can replicate each other; The potential limitation of energy conservation is equivalent to limiting the spectral radius.

\subsubsection{Optimization of hyper-parameters} \label{sec:Optimization of hyper-parameters}

Some parameters in ESNs can be adjusted in advance - three hyper-parameters $w^{in}$, $\alpha$ and $\rho(W_{res})$, which are initialized before training in basic ESN, leaky rate $a$, active function $f$, training readout methods and regularization parameters. Different initialization results in different performance.

Venayagamoorthy et al. \cite{DBLP:journals/nn/VenayagamoorthyS09} studied effects of spectral radius $\rho(W_{res})$ and settling time $ST$, the number of iterations of reservoir between input and output. The experiments showed that $\rho(W_{res})=0.8$ gives best result and the increase in ST adversely affects the ESN performance. Which showed that ESNs prefer short-term memory and desire no long-term echoing arrangement. 

The standard practice is the random initialization of the hyper-parameters, subject to few constraints. Although this results in a simple-to-solve optimization problem, it is in general suboptimal, and several additional criteria have been devised to improve its design. Rather than tuning by hand, more and more works focus on evolutionary optimization to optimize the parameters. 

The basic algorithms for searching optimal solution can apply in this context. Cai et al. \cite{DBLP:conf/iscid/CaiFWGZ18} have tested genetic algorithm (GA) and cross-validation (CV) method when optimizing the network parameters. The simulation showed that GA optimized ESN had better prediction accuracy while CV had better optimizing efficiency. Luo et al. \cite{DBLP:conf/iwcmc/LuoZSS20} applied PSO and made up two interacting aspects of regularization of $W_{out}$. Venayagamoorthy et al. \cite{DBLP:journals/nn/VenayagamoorthyS09} optimized relevant parameters of leaky-ESN by PSO. Martin et al. \cite{DBLP:journals/cin/MartinR15} combined self-assembly (SA) and PSO. Basterrech et al. \cite{DBLP:conf/isda/Basterrech13} tested the L2-Boost procedure \cite{B2003Boosting} for initializing the matrix spectrum when the network is large. Maat et al. \cite{DBLP:conf/ijcnn/MaatGP18} and Cerina et al. \cite{DBLP:conf/esann/CerinaFS19} optimized the ESN hyper-parameters by using Bayesian optimization by further reduce the optimization cost.

There are also some new methods for searching ESN hyper-parameters. Ishu et al. \cite{2004Identification} used a double evolutionary computation. The method first broad search the right parameters, then fine-tunes the the connectivity matrices of ESN. But it separated the topology and reservoir weights to reduce the search space. Thus, Ferreira et al. \cite{DBLP:conf/ijcnn/FerreiraL10} designed an evolutionary method for simultaneous optimization of parameters, topology and reservoir weights. Liu et al.\cite{DBLP:conf/iconac/LiuZ18} proposed a covariance matrix adaption evolutionary strategy ESN (CMA-ES-ESN) for searching $w^{in}, a$ and $\rho(W_{res})$. In \cite{DBLP:journals/access/BalaIISO20}, authors designed an improved grasshopper optimization algorithm (GOA) for the selection of ESN parameters. In \cite{DBLP:journals/sensors/HuangLC20}, an improved GA was proposed with the immigration strategy. In \cite{DBLP:conf/ijcnn/AkiyamaT19}, authors proposed a multi-step learning method for optimizing the hyper-parameters of multiple reservoirs ESNs. In \cite{DBLP:journals/nn/ThiedeP19}, Thiede et al. presented a gradient based optimization algorithm to minimize the error on specific tasks and specific structures of ESNs.

Exiting other different novel optimization methods, like improved fruit fly optimization algorithm (IFOA) \cite{DBLP:journals/ijon/ZhangQCL20}, fisher maximization based stochastic gradient descent (FM-SGD) \cite{DBLP:journals/ijon/OzturkCI20}, binary grey wolf algorithm (BGWO) \cite{DBLP:journals/ijon/LiuSLYCZ20a} and so on.

\subsubsection{Designs of regularization and training phase} \label{sec:Designs of regularization and training phase}

Many efficient training method have propoed. For example, Shutin et al. \cite{DBLP:journals/neco/ShutinZKP12} proposed variational Bayesian framework to train ESNs with automatic regularization and delay\&sum readout adaptation. In \cite{DBLP:conf/bica/GoudarziS14}, authors aimed to calculate the exact optimal output weights as a function of the structure of the system and the properties of the input and the target function. Scardapane et al. \cite{DBLP:journals/nn/ScardapaneWP16} decentralized algorithm in the case where data is distributed throughout a network of interconnected nodes. Chew et al. \cite{DBLP:journals/ar/ChewKN15} employed Tikhonov regularization to  train readouts and applied second-order proportional-integral-derivative feedback minimizes prediction bias. It reduced the number optimization parameters, while maintaining the estimation performance and following the excitation property of the estimator. Further, Couillet et al. \cite{DBLP:conf/ssp/CouilletWSA16} proposed a first theoretical performance analysis of the training phase of large dimensional linear ESNs based on random matrix theory.

The models discussed almost used supervised learning for training. Devert et al. \cite{DBLP:conf/ae/DevertBS07} adapted the ESN with unsupervised learning at the first time. They tickled the problem of unclear ground truth by adaptive (1+1)-evolution strategy with an optimisation task: optimising an ESN amounts to optimising a real-valued vector representing the plastic weights of the network. Similarly, Jiang et al. \cite{DBLP:conf/gecco/JiangBS08} applied the methods in evolutionary continuous parameter optimization and the double pole balancing control problem to the evolutionary learning of ESN. 

Many more efficient training schemes have been proposed to solve the ill-posed and over-fitting problem.

If the number of real-world training samples less than the size of the hidden layer, there may be an ill-posed problem. To overcome the ill-posed problem during the training process, Han et al. \cite{DBLP:journals/tnn/HanX18} employed the Laplacian eigenmaps to estimate the reservoir manifold and calculated output weights by the low-dimensional manifold. In \cite{DBLP:journals/nca/QiaoWY19}, authors replaced the current component by the newly generated network with more compact structure. Qiao et al. \cite{DBLP:journals/access/QiaoWYG18} developed an adaptive Levenberg–Marquardt algorithm to achieve convergence and stability. The calculation of the $W_{out}$ based on LM is equivalent to minimizing the objective function $E(W_{out})=\frac{1}{2}\sum_{q=1}^{Q}(\hat{y}_{q}-y_{q})^2$. Xu et al. \cite{DBLP:journals/tcyb/XuHQL19} proposed a hybrid regularized ESN when there are a large number of unknown output weights. The method employed a sparse regression with the $L_{\frac{1}{2}}$ regularization, having properties of unbiasedness and sparsity, and the $L_{2}$ regularization, having ability on shrinking the amplitude of the output weights. Meanwhile, the $L_{\frac{1}{2}}$ norm regularization term could also used to overcome the iterative numerical oscillation problem, described in \cite{DBLP:journals/eaai/LuXC20,DBLP:journals/tcas/LuoFSAY20}.

For the widely existed over-fitting issue, Yang et al. \cite{DBLP:journals/access/YangZAWQ18} proposed an incremental ESN (IESN) with an leave-one-out cross-validation (LOO-CV) error, which was used to automatically identify the network architecture and avoid over-fitting. For the same goal, in \cite{DBLP:conf/icann/NguyenKJK18}, Bayesian Ridge ESN (BRESN), having a regularization effect is proposed; In \cite{DBLP:journals/eaai/WangZWW19}, backtracking search optimization algorithm (BSA) determined the most appropriate output weights of ESN and avoid the over-fitted caused by the linear regression algorithm; In \cite{DBLP:journals/nn/YangQANW19}, the online sequential ESN with sparse recursive least squares (OSESN-SRLS) algorithm was proposed and two norm sparsity penalty constraints of output weights were separately employed to control the network size.

There are also some related studies about training phase. For example, good validation is key for good hyper-parameter tuning. Rather than using normal single validation split, Lukosevicius et al. \cite{DBLP:conf/icann/LukoseviciusU19} suggested k-fold cross-validation scheme, where the component that dominates the time complexity remained constant, not scaling up with k. Meanwhile, to provide a visualization of the modeled system dynamics, Lokse et al. \cite{DBLP:journals/cogcom/LokseBJ17} projected the output of the internal layer of ESNs on a lower dimensional space and enforced a regularization constraint that lead to better generalization capabilities. Further, Some works focused on how to make the results more accurate by changing the input more suitable for ESN. In \cite{DBLP:journals/tie/LiZO20}, a series of linear parameter-varying models was used to extract feature as the input of ESNs. Li et al. \cite{DBLP:conf/ijcnn/LiT20} combined ESNs with a preprocessing based on the Hodrick-Prescott (HP) filter. HP filter can extract different components from a single time-series data.

\section{Deep echo state networks} \label{sec:DeepESNs}

\begin{figure*}[t]
\centerline{\includegraphics[width=\linewidth]{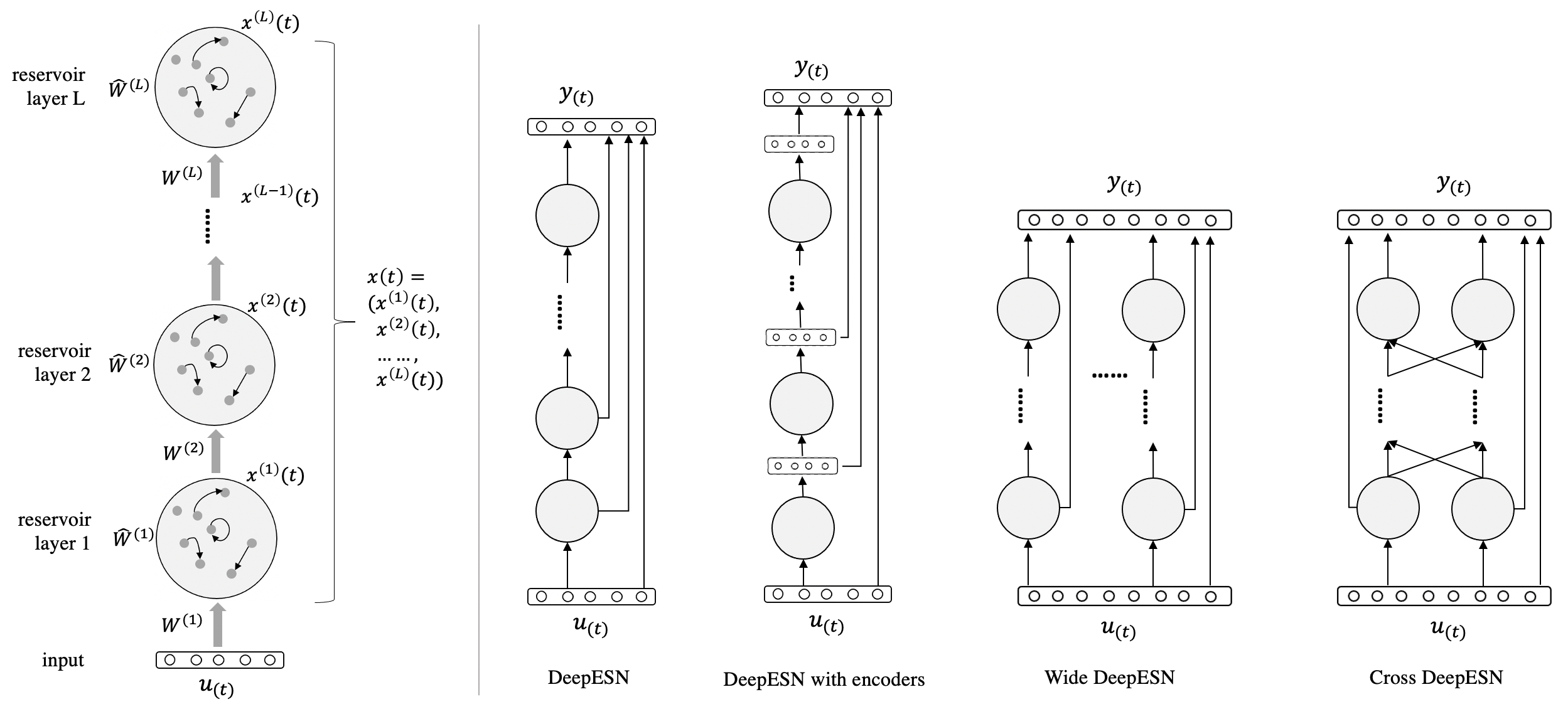}}
\caption{DeepESN and its variants}
\label{fig:DeepESN}
\end{figure*}

With the development of deep learning \cite{DeepLearningNature}, Stacking architecture \cite{DBLP:conf/icassp/DengYP12} has been introduced into reservoir computing. Deep echo state network (DeepESN) is the extension of ESNs to stacking multiple ESNs with deep learning framework. It was first introduced by Gallicchio et al. \cite{DBLP:conf/esann/GallicchioM16,Gallicchio2017Deep} in 2017. DeepESNs combines the advantages of both ESNs and deep learning. The former pursues conciseness and effectiveness, but the latter focuses on the capacity to learn abstract, complex features in the service of the task.

\subsection{Preliminaries} \label{sec:DeepESNs_Preliminaries}

\begin{definition}[Deep Echo State Networks (DeepESNs)] \label{def:DeepESNs}
DeepESN is a deep learning version of basic ESN. A DeepESN consists of an input layer, a dynamical stacking reservoirs component, and an output layer. The reservoir component of DeepESN is organized into a hierarchy of stacked recurrent layers, where the output of each layer acts as input for the next one.
\end{definition}

In a DeepESN, $u(t)\in R^{D}$ denotes the input value at time $t$ of the time series, $x^{(l)}(t)\in R^{N}$ denotes the state of the reservoir layer $i$ at time step $t$. Assuming $L$ is the number of hidden layers, $x(t)=\{x^{(l)}(t)|i=l,2,...L\} \in R^{D+LN}$. The state transition equation is in \ref{eq:DeepESN_trans}. Where $W^{(1)}\in R^{N\times D}$ is the input weight matrix. $W^{(l)}\in R^{N\times N}$ for $l>1$ is the weight matrix for inter-layer connections from $(l-1)$-th layer to $l$-th layer.  $\hat{W}^{(l)}\in R^{N\times N}$ is the recurrent weight matrix for layer $l$. $a^{l}\in [0,1]$ is the leaking rate for layer $l$. $f$ is a nonlinear function such as $\tan$ and sigmoid.

\begin{equation} \label{eq:DeepESN_trans}
\begin{aligned}
x^{(l)}(t)&=F(x^{(l-1)}(t),x^{(l)}(t-1))\\
&=(1-a^{(l)})x^{(l)}(t-1)+a^{(l)}f(W^{(l)}i^{(l)}(t)+\hat{W}^{(l)}x^{(l)}(t-1))\\
i^{l}(t)&=
\left\{ \begin{aligned} 
&u(t) \quad\quad l=1\\
&x^{l-1}(t) \quad l>1
\end{aligned} \right. \\
\end{aligned}
\end{equation}

In the basic DeepESN, the output turns to Equation \ref{eq:DeepESN_out}. The matrix $W_out \in R^{M(D+LN)}$ contains the feedforward weights between reservoir neurons and the $M$ output neurons.

\begin{equation} \label{eq:DeepESN_out}
y(t)=W_{out}x(t)=W_{out}(x^{(1)}(t),...,x^{(L)}(t))
\end{equation}

Figure \ref{fig:DeepESN} shows the structure of a DeepESN and its variants. As for the standard shallow ESN model, DeepESN can embed the input into a rich state representation by the stacking reservoirs. Thus, DeepESN can show the intrinsic properties of state dynamics in layered RNN architectures \cite{DBLP:conf/ijcnn/GallicchioM18}, enable a multiple time-scales representation of the temporal information, naturally ordered along the network’s hierarchy. DeepESN is also a efficiently trained deep neural network \cite{DBLP:journals/corr/abs-1712-04323}.

In \cite{DBLP:conf/esann/GallicchioM16}, authors used the experiment results of state entropy and memory to show that DeepESNs can enhance the effect of known ESN factors of network design. In \cite{DBLP:journals/corr/GallicchioMP17}, authors pointed out even in the simplified linear setting, progressively higher layers focus on progressively lower frequencies by applying means of frequency analysis to investigate the hierarchically structured state representation in DeepESNs. Above two studty showed that higher layers in the hierarchy can effectively develop progressively slower dynamics. 

Meanwhile, echo state property (ESP) have been generalized to DeepESNs in \cite{DBLP:journals/cogcom/GallicchioM17}. The authors gave the definition of a ESP based DeepESN in Equation \ref{def:DeepESP} and proofed that a sufficient condition and a necessary condition to hold in case of deep RNN architectures, shown in Equation \ref{eq:DeepESN_condition}. 

\begin{definition}[Echo State Property of DeepESNs (ESP of DeepESNs) ] \label{def:DeepESP} Assume a deepESN whose global dynamics are ruled by a function \ref{eq:DeepESN_trans}, Then the network has the echo state property if for each input sequence $U=[u(1),u(2),...u(N)]$ and all couples of initial states $x$, $x'$, it could hold the condition in Equation \ref{eq:DeepESP}. $\hat{F}(U,x)$ is the global state of the network which has started in the initial state $x$ and has been driven by the sequence $U$, 
\end{definition}

\begin{equation} \label{eq:DeepESP}
\begin{aligned}
&||\hat{F}(U,x)-\hat{F}(U,x')||\rightarrow 0 \quad \textit{as}\quad N \rightarrow \infty\\
&\hat{F}(U,x)=
\left\{ \begin{aligned} 
& x \quad if \quad U=\emptyset\\
& F((u(N),\hat{F}([u(1),...,u(N-1),x])) \quad if \quad U=[u(1),u(2),...u(N)]
\end{aligned} \right.\\
\end{aligned}
\end{equation}

\begin{equation} \label{eq:DeepESN_condition}
\begin{aligned}
&\textit{echo state property}\Rightarrow \rho_{g}=\max_{k=1,...,L}(\rho((1-a^{(k)})I+a^{(k)}\hat{W}^{(k)}))<1\\
&C=\max_{k=1,...,L}((1-a^{(k)})+a^{(i)}(C^{(i-1)}\overline{\sigma}(W^{i})+\overline{\sigma}(\hat{W}^{(i)})))<1  \Rightarrow \textit{echo state property}
\end{aligned}
\end{equation}

\subsection{Designs} \label{sec:DeepESNs_Designs}

Based on the basic DeepESN structure, many deformations have been propose.

Some existing works focused on the links between stacked reservoirs. In \cite{DBLP:journals/corr/abs-1711-05255, DBLP:journals/isci/MaSC20}, the deep structure was achieved along with encoders, shown in Figure \ref{fig:DeepESN}, DeepESN with encoders. An encoder projected the representations in a reservoir into a lower-dimensional space, and then the new representations were processed by the next reservoir. Thus, the structure was built by stacking projection layers and encoding layers alternately. In addition to the advantages of DeepESNs, this design could also attenuate the effects of the collinearity problem in ESNs. 

Like the idea of adding encoders, there are some other structures added different hidden layers between reservoirs. For example, Zhang et al. \cite{DBLP:journals/tfs/ZhangSWL0020} used fuzzy clustering between reservoir to reinforce embedding features for classification enhancement.  Arena et al. \cite{DBLP:journals/eaai/ArenaPS20} added reduction layer between reservoirs to avoid the influence of noise data. Bo et al. \cite{DBLP:journals/asc/BoWZ20} inserted some delayed modules between every two adjacent reservoirs. The reservoirs preserved the input characteristics by a relay mode and dealt with them asynchronously. This design achieved larger short-term memory capacity and richer dynamics.

Like the wide structure of multiple reservoirs in Figure \ref{fig:designs}, DeepESN can be organized by multiple modules of stacked reservoirs, shown in Figure \ref{fig:DeepESN}, Wide DeepESN. Further, the reservoirs in different modules could have the cross connections, shown in Figure \ref{fig:DeepESN}, Cross DeepESN. Based on the above connection structures, Carmichael et al. \cite{DBLP:journals/corr/abs-1908-08380,DBLP:journals/corr/abs-1808-00523} designed a DeepESN with modular architecture (Mod-DeepESN), which allowed for varying topologies of deep ESNs. 

Like the multiple input structure in Figure \ref{fig:designs}, DeepESN can also be designed to receive different components of data. In \cite{DBLP:journals/nca/KimK20}, the additive decomposition (AD) is applied to the time series as a preprocessor. Bianchi et al. \cite{DBLP:conf/esann/BianchiSLJ18} implemented a bidirectional reservoir, whose last state captures also past dependencies in the input. 

For the fundamental open issues of how to choose the number of layers in a deep RNN architecture, the work in \cite{DBLP:journals/nn/GallicchioMP18} proposed an automatic method for the design of DeepESNs by the analysis of differentiation of the filtering effects of successive. The proposed approach allows to tailor the DeepESN architecture to the characteristics of the input signals.

\section{Combinations of ESN and other models} \label{sec:xESNs}

\subsection{Combinations of ESN and basic machine learning models} \label{sec:ESN_ML}

Shi et al. \cite{DBLP:journals/tnn/ShiH07} combined ESN and support vector machine (SVM) as support vector echo-state machines (SVESMs). SVESMs performed linear support vector regression (SVR) in the high-dimension state space of reservoir. The new reservoir state variable of test data is generally written as Equation \ref{eq:SVESMs}. Where $\alpha_{s}(t)$ is the Lagrange multiplier corresponding to the reservoir state vector $x(t)$. $s$ is the number of support vectors. Scardapane et al. \cite{DBLP:journals/cogcom/ScardapaneU17} combined the standard ESN with a semi-supervised support vector machine (S$^{3}$VM) for training ESN's adaptable connections in semi-supervised learning setting.

\begin{equation} \label{eq:SVESMs}
\hat{x}^{test}(t)=\sum_{i=1}^{s}\alpha_{s}(t)x(t)^{T}x^{test}(t)
\end{equation}

Peng et al. \cite{DBLP:journals/nca/PengLLP14} combined ESNs and multiplicative seasonal autoregressive integrated moving average (ARIMA) model \cite{G1970Distribution} for mobile communication traffic series forecasting.

\begin{figure*}[t]
\centerline{\includegraphics[width=\linewidth]{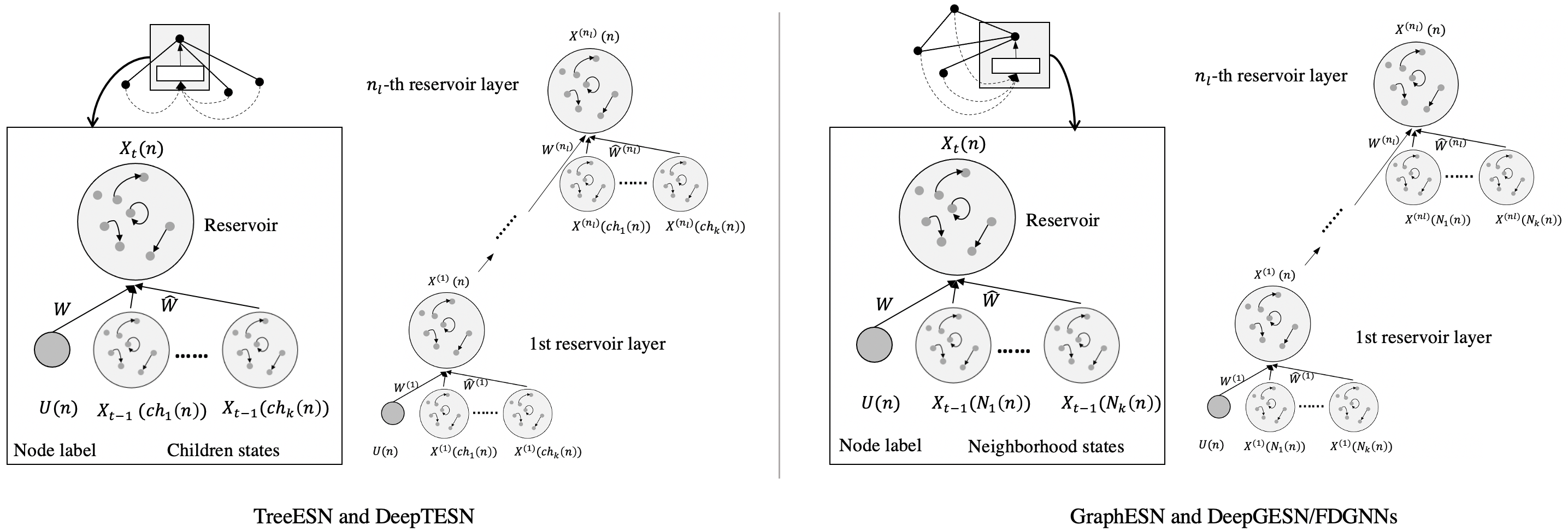}}
\caption{Tree echo state network and graph echo state network}
\label{fig:treegraphESN}
\end{figure*}

It is possible for ESN to handle tree-structure data, taking advantages from the compositionality of the structured representations both in terms of efficiency and in terms of effectiveness. The structure is in Figure \ref{fig:treegraphESN} left. Gallicchio et al. \cite{DBLP:journals/ijon/GallicchioM13} presented the tree ESN (TreeESN) model to extend the applicability of the ESN approach from sequence data to tree structured data. In TreeESN, the state for a vertex $n$ is computed from the input information attached to $n$ itself, and the states of $n$'s children nodes. Compared with the traditional ESN, the input of the children states takes the role of the state of $n$ in previous time-step. Then, in 2018, the authors showed the advanced version - deep TreeESN (DeepTESN) \cite{2018Deep}. Meanwhile, authors gave the theoretical properties, constraints, and empirical behavior of TreeESN and DeepTESN. Combining the deep learning, tree structure and reservoir computing take advantages from the layered architectural organization and from the compositionality of the structured representations both in terms of efficiency and in terms of effectiveness. The experimental results showed they outperformed the previous state-of-the-art in domains of document processing and computational biology.

It is also possible for ESN to handle graph-structure data. The structure is in Figure \ref{fig:treegraphESN} right. The concept of reservoirs operating on graph-structure data (GraphESN) has been first introduced in \cite{DBLP:conf/ijcnn/GallicchioM10}. GraphESN computed a state embedding for each vertex in an input graph. Similar to TreeESN, the state for a vertex $n$ is computed from the input information attached to $n$ itself, and the states of $n$'s neighbors. Further, in \cite{DBLP:conf/aaai/GallicchioM20}, the authors of TreeESN demonstrated that the DeepESN approach had good performance in the case of learning with graph. They designed Deep GraphESN, enabling the development of fast and deep graph neural networks (FDGNNs).

\subsection{Combinations of ESN and deep learning}\label{sec:ESN_DL}

\begin{figure*}[t]
\centerline{\includegraphics[width=\linewidth]{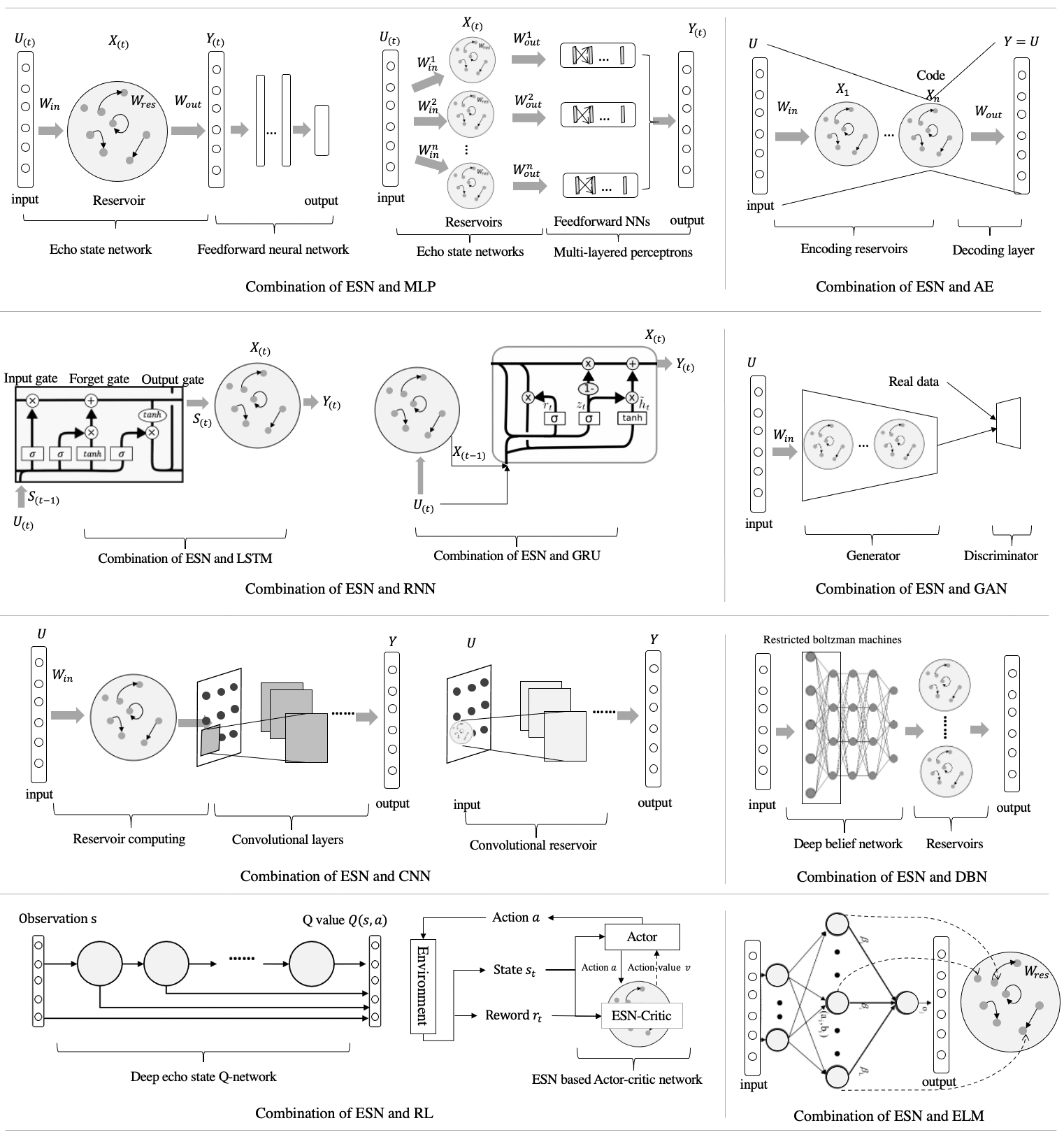}}
\caption{The combination of echo state network and other models}
\label{fig:XESN}
\end{figure*}

\begin{itemize}[leftmargin=10 pt]
    \item \textbf{Combinations of ESN and MLP}.
\end{itemize}

Babinec et al. \cite{DBLP:conf/icann/BabinecP06} combined ESN with feedforward neural networks. As shown in Figure \ref{fig:XESN} left, the output layer of echo state network is substituted with feedforward neural network and the one-step training algorithm, pseudoinverse matrix in Equation \ref{eq:ESN_solve}, is replaced with backpropagation of error learning algorithm. The approach was tested in temperature forecasting and had better performance than basic ESN and multi-layered perceptron (MLP). 

In \cite{DBLP:conf/esann/RodanT11}, this idea was extended to multi reservoirs structure, shown as Figure \ref{fig:XESN} right. Similarly, in \cite{DBLP:conf/isnn/WangPP11}, different reservoirs were assigned to different wavelet extracted from time series.

\begin{itemize}[leftmargin=10 pt]
    \item \textbf{Combinations of ESN and AE}.
\end{itemize}

Auto-encoder (AE) \cite{Vincent2008Extracting} structures are widely used in representation learning. It can extract the feature embedding in hidden layers with unsupervised learning technology. In \cite{DBLP:journals/isci/SunJYTLX19, DBLP:journals/kbs/WangWXWZ20, DBLP:journals/ijon/ChouikhiAHA19}, ESNs were used in AE structure, shown as Figure \ref{fig:XESN}. 

\begin{itemize}[leftmargin=10 pt]
    \item \textbf{Combinations of ESN and GAN}.
\end{itemize}

Generative adversarial networks (GAN) \cite{DBLP:conf/nips/GoodfellowPMXWOCB14}, consisting of two networks, a discriminator to distinguish between teacher data and generated samples and a generator to deceive the discriminator, have good performance to both generate new data and classifying data. In \cite{DBLP:conf/icann/AkiyamaT19}, authors combined ESN and GAN for generating samples that reflect the dynamics in the teacher data. The discriminator in their model was a feedforward neural network so that backpropagation could not be used through training. The structure is in Figure \ref{fig:XESN}.

\begin{itemize}[leftmargin=10 pt]
    \item \textbf{Combinations of ESN and RNN}.
\end{itemize}

Recurrent neural network (RNN) is a general term for a class of circularly linked neural networks. This type of neural network have shown excellent performance in processing sequence data. ESN is a special type of RNN. But at present, RNN refers to the deep learning network updated by error back-propagation. We use this expression in this section. In \cite{DBLP:journals/csr/LukoseviciusJ09}, authors compared RNN and ESN.
Traditional gradient-descent-based RNN training methods adapt all connection weights, including input-to-RNN, RNN-internal, and RNN-to-output weights. In ESN, only the RNN-to-output weights are adapted.

RNNs were considered difficult to train, as their memory capabilities are often thwarted in practice by the exploding/vanishing gradients problem. As an attempt to solve these problems, long short-term memory cells (LSTM) \cite{DBLP:journals/neco/HochreiterS97} were designed to selectively forget information about old states and pay attention to new inputs. In \cite{DBLP:conf/icann/PopovKSO19}, BiESN and BiLSTM were compared in the task of word sense disambiguation. The experiments confirmed the ability of the BiESN to capture more context information. And the main advantage of ESN over LSTM is the smaller number of trainable parameters and a simpler training algorithm. However, in ESN, the simplicity of not training its hidden layer may restrict reaching a better performance. In \cite{DBLP:conf/ciarp/LopezVAG17}, the authors combined LSTM and ESN, replacing the hidden units of ESN by LSTM blocks. Initially all weights were randomly initialized and only the weights of output layer are learned. Then, whole network was trained by a gradient method but fixing the output weights. Yang et al. \cite{DBLP:journals/tim/YangZL20} also combined ESN and LSTM by applying the gate idea of LSTM to the reservoirs of ESN, aiming at the problem that information cannot be effectively retained, shown in Figure \ref{fig:XESN}.

Gated recurrent unit (GRU) \cite{DBLP:conf/emnlp/ChoMGBBSB14} is another enhanced version of RNN. In \cite{DBLP:conf/ijcnn/WangJH20, DBLP:conf/inista/SarliGM20}, authors presented GRU-based ESNs to tackle complex real-world tasks while reducing the computational costs. The reservoir unit was replaced by the sparsely connected GRU neurons. Besides, the gating mechanisms improved the network’s ability to deal with long-term dependencies within the data.

In addition to the direct combination of the two models, some works have incorporated the idea of RNN into ESN. Zheng et al. \cite{DBLP:journals/access/ZhengQLXZM20} found that in ESNs, no explicit mechanism captures the multi-scale characteristics of time series. They proposed long-short term echo state networks (LS-ESNs) consisting of three independent reservoirs. Each reservoir has recurrent connections of a specific time-scale to model the temporal dependencies of time series. Three state transition equations are in Equation \ref{eq:LS-ESN}.

\begin{equation} \label{eq:LS-ESN}
\begin{aligned}
&x_{typical}(t)=(1-a)x_{typical}(t)+a f(W_{in}u(t)+W_{res}x_{typical}(t-1))\\
&x_{long}(t)=(1-a)x_{long}(t)+a f(W_{in}u(t)+W_{res}x_{long}(t-k))\\
&x_{short}(t)=(1-a)x_{short}(t-1)+a f(W_{in}u(t)+W_{res}x_{short}(t-1)) \quad until \quad x_{short}(t-m+1)
\end{aligned}
\end{equation}

Li et al. \cite{DBLP:conf/ijcnn/LiT20a} also found that ESN-based models have used only single-span features. Thus, they proposed multi-span DeepESN for features for considering the relations on different scales of a sequence. The state transition equation is in Equation \ref{eq:span-ESN}. Where the input data is at the time $t$ in the $g$-th group of the $l$-th reservoir layer. Group is a subsequence of raw sequence with a specific time span. 

\begin{equation} \label{eq:span-ESN}
x^{(l)}_{(g)}(t)=(1-a^{(l)}_{(g)})x^{(l)}_{(g)}(t)+a^{(l)}_{(g)} f(W^{(l)}_{in(g)}u(t)+W^{(l)}_{res(g)}x^{(l)}_{(g)}(t-1))
\end{equation}


\begin{itemize}[leftmargin=10 pt]
    \item \textbf{Combinations of ESN and CNN}.
\end{itemize}

In recent years, convolutional neural networks (CNNs) have achieved great success in end-to-end pattern recognition. Many CNN-based models, such as VGG \cite{DBLP:journals/corr/SimonyanZ14a}, ResNet \cite{DBLP:conf/cvpr/HeZRS16} have been widely proposed. To combine the advantages of both ESN and CNN, in \cite{DBLP:journals/access/ZhangZL19}. for series classification tasks, feature extraction was implemented by ESN, and the classification was obtained by the CNN. The connection of this structure is shown in Figure \ref{fig:XESN}. Which had the advantage of the long-and-short term memory brought by the echo state and high-accuracy performance brought by the deep CNN.

In addition to connecting ESN and CNN directly, in \cite{DBLP:conf/robio/ZhangCLZ18}, convolutional dynamics were built by extending randomly connected reservoirs to convolutional structures in the input-to-state transition, shown as Figure \ref{fig:XESN}.The transition function of basic ESN changed to Equation \ref{eq:CNNESN}. Where $(W_{res})^{l},l=1,...L$ is a convolution at delay $l$ and $L$ is the length of time delays. This well optimized convolutional ESN can establish long-term time series predictions by ESN and short-term dependence by building the convolutional architecture.

\begin{equation} \label{eq:CNNESN}
x(t)=f(W_{in}u(t)+\sum_{l=1}^{L}(W_{res})^{l}\otimes x(t-1)+W_{back}y(t))
\end{equation}

\begin{itemize}[leftmargin=10 pt]
    \item \textbf{Combinations of ESN and DBN}.
\end{itemize}

Restricted Boltzmann machine (RBM) \cite{DBLP:journals/cogsci/AckleyHS85,DBLP:conf/nips/TehH00,DBLP:reference/ml/Hinton17} is the first multi-layer learning machine inspired by statistical mechanics. The adjustment of weights and the state evolution are determined by a certain probability distribution. In \cite{DBLP:journals/tr/FinkZW15}, authors proposed a fuzzy classification approach applying a combination of supervised ESN and unsupervised RBM, achieving two main strengths - hidden features extraction by RBM and dynamic patterns learning by ESN. The structure was the is to link two structures directly, the input of ESN is the output of RBM. In \cite{DBLP:journals/iotj/SunMLWLWG20}, an enhanced echo-state RBM (eERBM) was presented for network traffic prediction.

Deep belief network (DBN) is a typical unsupervised learning algorithm developed by Hinton et al. \cite{DBLP:journals/neco/HintonOT06, DBLP:reference/ml/Hinton10a}. DBN consisting of layers of RBM has strong ability to extract intrinsic nonlinear features of data through its hierarchical structure. Many works \cite{DBLP:journals/ria/ChengZ19, DBLP:journals/kbs/SunLLHL17} have explored ways to combine the advantages of DBN and ESN. Most of them is also the direct connection of them, show in Figure \ref{fig:XESN}.

\begin{itemize}[leftmargin=10 pt]
    \item \textbf{Combinations of ESN and RL}.
\end{itemize}

Reinforcement learning (RL) has been shown to be widely successful in many applications, ranging from playing games \cite{DBLP:journals/corr/MnihKSGAWR13, DBLP:journals/nature/SilverHMGSDSAPL16} to robotics \cite{DBLP:conf/icra/FinnL17, DBLP:conf/aaai/WhitmanBTC20}. RL algorithms can be divided into three categories - value-based, policy-based and actor-critic.

Deep Q-network (DQN) \cite{DBLP:journals/nature/SilverHMGSDSAPL16, DBLP:journals/nature/MnihKSRVBGRFOPB15,DBLP:conf/aaai/HasseltGS16}, one of the common value-based algorithms, has only one value function network. DeepESNs have abilities to replace the deep neural networks in DQN and compute Q-value \cite{DBLP:conf/icann/SzitaGL06,DBLP:journals/corr/abs-2010-05449} with faster computing speed and global optimal solution.

Actor-critic network (AC) \cite{DBLP:conf/icml/MnihBMGLHSK16} is actually a combination of policy-based and value-based RL methods. Existing some works \cite{DBLP:conf/esann/Oubbati11,DBLP:conf/iros/SchmidtBP14,DBLP:conf/rita/MatsukiS17,DBLP:conf/smc/Koprinkova-HristovaOP10} studied the application of ESN in AC. Most of them put ESN into Critic net. The main advantage is that ESN can learn a “memory task” through RL with back propagation \cite{DBLP:conf/iros/SchmidtBP14}. Some tasks, like dynamic programming, need fast online training algorithms and robust modelisation of robot-environment interaction. Most of the method embeded RNN in RL networks for exhibiting continuous dynamics. To achieve fast online training and overcome the training difficulties of RNNs, ESN has been used in \cite{DBLP:conf/esann/Oubbati11}. 

\begin{itemize}[leftmargin=10 pt]
    \item \textbf{Combinations of ESN and ELM}.
\end{itemize}

Extreme learning machine (ELM) \cite{DBLP:journals/ijon/HuangZS06, DBLP:journals/air/DingZZXN15} is a single hidden layer feedforward neural network which randomly chooses hidden nodes and analytically determines the output weights. Like ESN, the most significant characteristic of the ELM is that it provides good generalization performance at extremely fast learning speed than gradient-based learning algorithms \cite{DBLP:journals/ijon/HuangZS06}. Or In other words, ESN is a sort of ELM designed for time series. Some works \cite{DBLP:conf/iconip/SiqueiraBAL12,DBLP:conf/ciasg/JayawardeneV14, DBLP:conf/ijcnn/CarvalhoSFB18,DBLP:journals/eaai/RibeiroRS20} have used ESN and ELM in real-time study application, like seismic multiple removal \cite{DBLP:conf/ijcnn/CarvalhoSFB18} and photovoltaic power prediction \cite{DBLP:conf/ciasg/JayawardeneV14}. where the fast computation is required. In \cite{DBLP:journals/asc/Ertugrul20}, An ESN-based ELM could model a non-sequential dataset or system with a high generalization capacity and no requirement of any optimization stage. The structure is shown in Figure \ref{fig:XESN}.

\section{Applications} \label{sec:Applications}

For what regards the experimental analysis in applications, ESNs were shown to bring several advantages in both cases of synthetic and real-world tasks. 

\subsection{Benchmark datasets} \label{sec:Benchmark}

\begin{itemize}[leftmargin=10 pt]
    \item \textbf{Mackey-Glass System}.
\end{itemize}

Mackey-Glass (MG) system  is a classical chaotic system that is often used evaluating time series prediction models. A standard discrete time approximation to the MG delay differential equation is in Equation \ref{eq:MGS}

\begin{equation} \label{eq:MGS}
y(t+1)=y(t)+\delta(a\frac{y(t-\frac{\tau}{\delta})}{1+y(t-\frac{\tau}{\delta})^{n}}-by(t))
\end{equation}

Where the parameters $\delta,a,b,n$ are usually set to $0.1, 0.2, -0.1, 10$. The system becomes chaotic if $\tau > 16.8$ \cite{Jaeger2004Harnessing}, most existing works \cite{DBLP:journals/isci/MaSC20} set $\tau=17$ and initial $y(0)=1.2$. The task can be predict the value in time series after step k. Which means using the data $u(1,...,t)$ to forcast the value of $u(t+k)$.

\begin{itemize}[leftmargin=10 pt]
    \item \textbf{Multiple Superimposed Oscillator series}.
\end{itemize}

Multiple Superimposed Oscillator Series (MSO) time series prediction problem is also a benchmark, the MSO time series data are generated by summing up several different frequency sine wave functions \cite{Xue2007Decoupled}, which is derived as shown in Equation \ref{eq:MSO}:

\begin{equation} \label{eq:MSO}
y(t)=\sum_{i=1}^{m}sin(\alpha_{i}t)
\end{equation}

n general, the MSO can also be written as MSOm, where $m$ represents the number of summed sine waves. The forecast difficulty is increased with the increase in the number of summed sine waves. To test ESN, most existing works set $s=5$ and $\alpha_{1} =0.2, \alpha_{2}=0.311, \alpha_{3}=0.42, \alpha_{4}=0.51, \alpha_{5}=0.63$. \cite{DBLP:journals/tnn/QiaoLHL17,DBLP:conf/ijcnn/WangJH20}

\begin{itemize}[leftmargin=10 pt]
    \item \textbf{Lorenz system}.
\end{itemize}

The Lorenz system is one of the most typical benchmark tasks for time series prediction, described in Equation \ref{eq:Lorenz}:

\begin{equation} \label{eq:Lorenz}
\left\{\begin{aligned}
&\frac{dx}{dt}=-ax(t)+ay(t)\\
&\frac{dy}{dt}=bx(t)-y(t)-x(t)z(t)\\
&\frac{dz}{dt}=x(t)y(t)-cz(t)\\
\end{aligned}\right.
\end{equation}

Where $4a=10$, $b=28$, and $c=8/3$. For getiting Lorenz time series, the Runge-Kutta method can generate sample of length $L$ from the initial conditions ($x_{0}$,$y_{0}$,$t_{0}$) with a step size of 0.01. For prediction task, methods can predict the data in x-axis, y-axis or z-axis.

\begin{itemize}[leftmargin=10 pt]
    \item \textbf{Rossler system}.
\end{itemize}

Rossler system a classical system which sufficiently chaotic phenomenon. It can be expressed in Equation \ref{eq:Rossler}  

\begin{equation} \label{eq:Rossler}
\left\{\begin{aligned}
&\frac{dx}{dy}=-y-z\\
&\frac{dy}{dz}=x+ay\\
&\frac{dz}{dt}=b+z(x-c)\\
\end{aligned}\right.
\end{equation}

Where $x, y, z$ are system variables, $a, b, c$ are adjustable coefficients, and t denotes time dimension. When a = 0.2, b = 0.2, c = 5.7, Equation \ref{eq:Rossler} produces chaos. For prediction task, methods can predict the data in x-axis, y-axis or z-axis.

\begin{itemize}[leftmargin=10 pt]
    \item \textbf{NARMA system}.
\end{itemize}

Nonlinear auto-regressive moving average system (NARMA) is a highly nonlinear system incorporating memory. The tenth-order NARMA system depends on outputs and inputs 9 time steps back, which is difficult to identify. The tenth-order NARMA system identification problem is often used to test ESN models. The system is described in Equation 

\begin{equation} \label{eq:NARMA}
y(t+1)=0.3y(t)+0.05y(t)\sum_{i=0}^{9}y(t-i)+1.5u(t-9)u(t)+0.1
\end{equation}

where $u(t)$ is a random input of the system at time step $t$, drawn from a uniform distribution over [0, 0.5]. The output $y(t)$ is initialized by zeros for the first ten steps $(t = 1,...,10)$. One-step-ahead direct prediction on this time series is usually the task for ESN.

\begin{itemize}[leftmargin=10 pt]
    \item \textbf{Sunspot datasets}.
\end{itemize}

The smoothed monthly mean sunspot numbers, which are one of the typical time series, are usually used as the benchmark prediction problems. The sunspot number is a dynamic manifestation of the strong magnetic field in the Sun’s outer regions. Due to the complexity of potential solar activity and high non-linearity of the sunspot series, analyzing these series is a challenging task. There are two public datasets - SILSO and NOAA. The task can be forecasting the sunspot number in the future.

The World Data Center, Sunspot Index and Long-term Solar Observations (SILSO) \cite{SILSO}, provides an open-source monthly sunspot series from 1749 to 2020. 

Another dataset comes from the National Oceanic and Atmospheric Administration (NOAA) \cite{NOAA}, which includes 3174 samples of sunspot activity numbers from 1750 to 2013. 

\begin{itemize}[leftmargin=10 pt]
    \item \textbf{Temperature datasets}.
\end{itemize}

A temperature series can show chaotic behavior, which makes them difficult to be forecasted accurately. There are two public benchmark temperature datasets - USHCN and DMTM. To test ESN model, the task can be climate forecasting after k days on these two datasets.

The U.S. Historical Climatology Network Monthly dataset (USHCN) \cite{USHCN} is publicly available and consists of daily meteorological data of stations in 48 states of the United States spanning from 1887 to 2014. It has everyday climate variables for each station, temperature, precipitation and snow precipitation. 

Another temperature dataset is daily minimum temperatures in Melbourne (DMTM). There are 3650 minimum temperature points in Melbourne collected from January 1st, 1981 to December 31th, 1990.

\begin{itemize}[leftmargin=10 pt]
    \item \textbf{Medical datasets}.
\end{itemize}

The Medical Information Mart for Intensive Care - three version dataset (MIMIC-III) \cite{mimic-iii} is a public dataset collected at Beth Israel Deaconess Medical Center from 2001 to 2012. It contains over 58,000 hospital admission records of 38,645 adults and 7,875 neonates. Potential tasks include mortality prediction, disease prediction, vital sign control, and patient subtyping.

\begin{itemize}[leftmargin=10 pt]
    \item \textbf{Others}.
\end{itemize}

Henon map system \cite{M1976A}: a typical discrete-time dynamical sys-tem exhibiting a characteristic chaotic behavior;
Labour dataset \cite{Labour}: Monthly unemployment rate in US from 1948 to 1977;
Gasoline dataset \cite{gasoline}: US finished motor gasoline product supplied;
Santa Fe laser series \cite{Santa}: the intensity of a NH3-FIR laser in a chaotic regime;
Electricity dataset \cite{DBLP:journals/jors/Taylor03}: Half-hourly electricity demand in England;
Photovoltaic power dataset \cite{Solarpower}: Solar power data in Twentynine Palms and Whidbey Island;
UCR database archive \cite{UCRArchive}: 358 publicly available synthetic and real time series databases.

\subsection{Abnormal data orientated} \label{sec:abnormal_data}

In practice, the uncertainty in real-world dynamic system is quite prevalent such as data noise, data imbalance and multiple data formats. With the unclear data, the low accuracy of ESNs for real-world tasks showed up.

\begin{itemize}[leftmargin=10 pt]
    \item \textbf{Data noises}.
\end{itemize}

Data noise is the most common problem. For example, the various noises are often introduced into an industrial system by sensors. The most intuitive method \cite{Xue2007Decoupled,2007Online} addressed the problem by data preprocessing in some specific fields such as nonlinear system modeling. For example, Xu et al. \cite{DBLP:journals/isci/XuHL18} gave a wavelet-denoising algorithm. Some works tickled the issue according to concerning the output uncertainty and internal states uncertainty causing by data noise. For example, in \cite{DBLP:journals/ijon/ShengZLW12}, noise addition was integrated into ESN to describe the additive noises of uncertainty of internal states the output, shown in Equation \ref{eq:noise}, where $v$ and $n$ are independent white Gaussian noise sequences. Some works used limitation in objective to alleviate the influence from noises. For example, Senn et al. \cite{DBLP:conf/annpr/SennK20} proposed abstract learning with the constrain $|W_{out}X_{r}|\leq Y_{r}$ of the objective function for the noise data, where $Y_{r}$ is the maximum deviations could be tolerate and $X_{r}$ is distance around the center value of $X$. Then Rigamonti et al. \cite{DBLP:journals/ijon/RigamontiBZRGP18} gave the adaptive solution. 

\begin{equation} \label{eq:noise}
\begin{aligned}
x(t)&=f (W_{in}u(t)+W_{res}x(t-1))+v_{t-1}\\
y(t) &= W_{out}x(t)+n_{t}   
\end{aligned}
\end{equation}

For outliers data, in \cite{DBLP:conf/ifsa/LiuZW13,DBLP:journals/sj/YangZ20}, authors used Gaussian process to avoid the accumulative iteration error by establish the direct relation between the reservoir state and outputs. In \cite{DBLP:journals/tnn/LiHW12}, authors replaced the Gaussian distribution with Laplace distribution. In \cite{DBLP:journals/asc/ShenCZYJ18}, Shen et al. used variational inference methods. In \cite{DBLP:conf/ijcnn/ZhangGWC18}, Generalized correntropy induced metric (GCIM) was robust to outliers with a proper shape parameter. And Guo et al. \cite{DBLP:journals/ijon/GuoWCX17} gave a correntropy induced loss function for ESN to improve its anti-noise capacity.

\begin{itemize}[leftmargin=10 pt]
    \item \textbf{Imbalanced and incomplete data}.
\end{itemize}

Many actual datasets are mainly imbalanced. For example, in electronic health records (EHRs), the records of common diseases are much more than those of rare diseases. And for classification task, the lack of data in a minority class may lead to uneven accuracy. Chen et al. \cite{DBLP:journals/nca/ChenZHHS20} used well-trained ESN as the memory to replace the method of setting invariable thresholds between different classes.

Meanwhile, in the complex industrial environment, data missing often occurred in the process of data acquisition and transition, causing the incomplete dataset. The proposal of a deep bidirectional ESN (DBESN) \cite{DBLP:journals/access/WangWLZW19} could model the incomplete dataset. It extracted the features along with forward and backward time scales and used deep auto-encoder to construct the incomplete output and input.

\begin{itemize}[leftmargin=10 pt]
    \item \textbf{Dynamic data format}.
\end{itemize}

Under normal conditions, the inputs of existing ESNs are mostly limited to static and not dynamic temporal patterns.

For process dynamic data, Tanisaro et al. \cite{DBLP:conf/icmla/TanisaroH16} used the idea of time warping invariant neural network (TWINN) \cite{1992Time}. Where the time warping in networks can be considered as a variation of the speed of the process. 
Zhang et al. \cite{DBLP:journals/ijista/Song-linX17} noticed the timeliness of data (time varying data) and proposed adaptive forgetting (AF) factor to ESN. AF could automatically tune the valid data window size according to prediction error magnitude. Tanaka et al. \cite{DBLP:journals/ajiips/TanakaNH19} addressed data that were generated by multiple dynamical systems switched with time by using time-varying reservoir. The reservoir adaptively changed depending on a statistical property of teacher data.  For a class of discrete-time dynamic nonlinear systems (DDNS), Yao et al. \cite{DBLP:journals/nn/YaoWZ19} adjusted reservoir state adaptively according to the characteristics of input signals. Moreover, in dynamic data form, the reliability of former data also needs to be considered. In \cite{DBLP:journals/nca/HuangQZWS18}, the authors presented a recursive Bayesian linear regression ESN (RBLR-ESN) to change the confidence level of the prior data. The model was well-designed for long-range sequence.

The relation in time series data is mainly one-way time sequence, in natural language processing context, the relations reflect in bidirectional sequence. Schaetti \cite{DBLP:conf/clef/Schaetti18b} described Bidirectional ESN to model the order reverse order relations in sentences.

\begin{itemize}[leftmargin=10 pt]
    \item \textbf{Expert knowledge}.
\end{itemize}

For a specific system, ESN may achieve better performance when incorporating the expert knowledge. For example, a physics-informed ESN \cite{DBLP:conf/iccS/DoanPM19,DBLP:journals/corr/abs-2011-06769} could ensure that their predictions do not violate physical laws compared to conventional ESNs.

\begin{itemize}[leftmargin=10 pt]
    \item \textbf{High dimensional data}.
\end{itemize}

Besides sequence data, ESNs can also applied in multi-dimensional data, such as those observed in the context of image recognition (2D imamge), renewable energy (3D wind modeling) and human centered computing (3-D inertial body sensors). 

For image processing, in \cite{DBLP:conf/cse/SchaettiSC16}, ESNs were applied in classifying the digits of the MNIST, a basic database for computer vision; Wen et al. \cite{DBLP:conf/acii/WenLL15} designed an ensemble convolutional ESN for face recognition; And Duan et al. \cite{DBLP:journals/tnn/DuanW16} used ESN for image restoration. For video data, in \cite{DBLP:conf/esann/MiciHW16}, ESNs were presented for the task of classifying high-level human activities. To cater for 3-D and 4-D processes, Quaternion-valued ESN (QESNs) \cite{DBLP:journals/tnn/XiaJM15} used quaternion nonlinear activation functions with local analytic properties on quaternion variable $q=q_{r}+\iota q_{\iota}+jq_{j}+\kappa q_{\kappa}$.

\subsection{Real-world tasks orientated} \label{sec:real-world_tasks}

As pertains to real-world problems, the ESN approaches recently proved effective in a variety of domains.

\begin{itemize}[leftmargin=10 pt]
    \item \textbf{Industrial applications}.
\end{itemize}

ESNs are widely used in industry \cite{DBLP:conf/ifip12/DettoriMCS20,DBLP:conf/eann/CollaMDCM19,DBLP:conf/ijcnn/MansoorGM20,DBLP:conf/ifip12/DettoriMCS20}, considering in particular the energy and manufacturing sectors.
in energy fields, ESNs have applied in traditional energy forecasting \cite{energy}, like prognostic of fuel cells \cite{DBLP:journals/mcs/MorandoJHGZ17,DBLP:journals/tie/LiZO20,DBLP:conf/iecon/MorandoJGZH13}, health of batteries \cite{DBLP:journals/sensors/SanchezAOC18}, oil production platform control \cite{DBLP:journals/eaai/JordanouAC19}, gas prediction \cite{DBLP:journals/tim/YangZL20} and electric ship \cite{DBLP:conf/ijcnn/DaiVH09}, and clean energy forecasting, like wind power generation\cite{DBLP:conf/cies/AquinoNSLCLF14,DBLP:conf/ideal/Dorado-MorenoGS18} and photovoltaic power prediction \cite{DBLP:conf/ciasg/JayawardeneV14,DBLP:conf/iciai/ZhangLCYZL19,DBLP:conf/iciai/ZhangLCYZL19,DBLP:journals/nn/YaoWZ19}. ESN-based models have also been used in manufacturing, such as motor control \cite{DBLP:conf/icra/SalmenP05}, detection and diagnosis of anomaly and fault\cite{DBLP:conf/ipsn/ObstWP08,DBLP:conf/dcoss/ChangTB09,DBLP:conf/ijcnn/WoottonDH18,DBLP:journals/tii/LongZL20,DBLP:journals/tfs/ZhangSWL0020,DBLP:journals/eaai/XuBAZ20,DBLP:journals/eaai/LuXC20,DBLP:journals/access/BalaIISO20}, production system monitor \cite{DBLP:journals/nn/Venayagamoorthy07,DBLP:journals/nn/AntoneloCF17} and communications of sensors \cite{DBLP:conf/wises/MathewsP08,DBLP:conf/isnn/QinHZ09,DBLP:books/sp/09/KrauseBDS09,DBLP:conf/isda/SongFKL09,DBLP:conf/rose/HarischandraD12,DBLP:conf/ijcnn/KuwabaraNKBGCP12}, mobile \cite{DBLP:journals/nca/PengLLP14,DBLP:journals/nn/BianchiSURS15}, satellite \cite{DBLP:conf/vtc/BauduinSMH15}, wireless \cite{DBLP:journals/iet-com/GideonNT17} and 5G Systems \cite{DBLP:journals/esticas/BaiYZJL20}.

\begin{itemize}[leftmargin=10 pt]
    \item \textbf{Medical applications}.
\end{itemize}

Medical field is also an important application of ESNs. At present, the main practices include feature learning of vital signs, such as electroencephalogram (EEG)-based feature extraction \cite{DBLP:journals/isci/SunJYTLX19,DBLP:journals/cbm/KimJ19,DBLP:conf/ijcnn/FouratiAJA20} and event detection \cite{DBLP:conf/embc/AyyagariJW14,DBLP:conf/embc/AyyagariJW15,DBLP:conf/iconip/FouratiAASA17,DBLP:journals/nca/RenDW19,DBLP:journals/nca/RenDW20}, electrocardiogram (ECG)-based feature clustering \cite{DBLP:conf/icann/RuizGJM20}, atrial fibrillation detection \cite{DBLP:conf/idaacs/PetrenasML11,DBLP:journals/tbe/PetrenasMSL12}, arterial blood pressure prediction \cite{DBLP:conf/amia/FongMRR14} and magnetic resonance imaging (MRI)-based disease diagnosis \cite{DBLP:conf/mibam/WismullerDAWHN15}. 
Meanwhile, there are many methods for disease diagnosis. For example, ESNs methods have been applied in the diagnosis of parkinson’sdisease \cite{DBLP:conf/esann/GallicchioMP18,DBLP:journals/artmed/LacySL18}, prediction of dialysis in critically ill patients \cite{DBLP:journals/midm/VerplanckeLSBTMD10}, diagnosis of breast cancer \cite{DBLP:conf/cicare/WajidHL14}, classification of ICU sepsis \cite{DBLP:conf/cinc/AlfarasVG19}, detection of oral cancer \cite{DBLP:journals/jms/Al-MaaitahA18}, classification of bone marrow cell \cite{DBLP:journals/nca/KainzBAA17} and prediction of blood glucose concentration for type 1 diabetes \cite{DBLP:journals/ijon/LiTWW20}

\begin{itemize}[leftmargin=10 pt]
    \item \textbf{Spatio-temporal applications}.
\end{itemize}

Many works \cite{DBLP:journals/tnn/SohD15,DBLP:conf/icra/SchaffernichtBL17,DBLP:journals/corr/abs-1806-10728} have designed ESN-based model for processing spatio-temporal data. Spatio-temporal data widely appear in traffic application. In this domain, ESNs can be used in traffic forecasting \cite{DBLP:journals/corr/abs-2004-08170,DBLP:conf/icnc/AnSZ11,DBLP:conf/itsc/SerLBV19,DBLP:journals/corr/abs-2004-08170}, destination prediction \cite{DBLP:journals/ijon/SongWS20}, bike-sharing application \cite{0Deep,DBLP:journals/nca/KimK20} and pedestrian counting system \cite{DBLP:journals/ejes/MathewsP09}. 

There are a lot of spatio-temporal data in astronomy and meteorology. sIn the field of astronomy, sunspot activity forecasting is a hot issue and the sunspot datasets is one of the benchmark dataset as we described in \ref{sec:Benchmark}. Many works tested their models on this task \cite{DBLP:conf/esann/SchwenkerL09}. Besides, there are also many ESN-based models designed for solar irradiance prediction \cite{DBLP:conf/isda/BasterrechZPM13,DBLP:journals/asc/LiWLFL20,DBLP:conf/ecc/BasterrechB14,DBLP:conf/eann/KmetK15,DBLP:journals/tii/WuLX21}, metereological forecasting\cite{0Deep,DBLP:journals/nca/KimK20,DBLP:journals/tnn/XuYHQL19}, temperature prediction \cite{DBLP:journals/nca/HuangQZWS18}, water/stream flow forecasting \cite{DBLP:conf/ijcnn/SacchiOPCS07,DBLP:conf/ideal/SiqueiraBAF12,DBLP:conf/iconip/SiqueiraBAL12} and wind speed Forecasting\cite{DBLP:conf/ijcnn/Qiao09,DBLP:conf/ciarp/LopezVAG17,DBLP:conf/amcc/ChitsazanFNT17}

\begin{itemize}[leftmargin=10 pt]
    \item \textbf{Financial applications}.
\end{itemize}

Financial forecasting is the focus of time series forecasting \cite{0Deep,DBLP:journals/nca/KimK20}, using ESNs model, many methods have proposed for stock data mining \cite{DBLP:conf/pakdd/LinYS08}, stock price prediction \cite{DBLP:journals/eswa/LinYS09}, stock trading system control \cite{DBLP:journals/eswa/LinYS11} and financial Data Forecasting \cite{DBLP:conf/iconip/LiuSLFCZD18}.

\begin{itemize}[leftmargin=10 pt]
    \item \textbf{Computer vision}.
\end{itemize}

In the field of computer vision (CV), ESN have been used in image processing, such as image segmentation \cite{DBLP:conf/ijcnn/SouahliaBBC16,DBLP:journals/cogcom/MeftahLB16,DBLP:conf/icaci/SouahliaBBC17,DBLP:conf/mesas/DonkorSB18,DBLP:journals/concurrency/SouahliaBBABC20}, image restoration \cite{DBLP:journals/tnn/DuanW16}, facial expression recognition \cite{DBLP:conf/acii/WenLL15}. Many methods also could process radio audio data \cite{DBLP:conf/isnn/SquartiniCRP07,DBLP:journals/nn/GallicchioMP18, 0Deep, DBLP:conf/esann/GallicchioMP19}, like video traffic \cite{DBLP:conf/ccis/SunCLCL12}, video annotation \cite{DBLP:conf/icmla/RoychowdhuryM18}, audio Classification \cite{DBLP:journals/cogcom/ScardapaneU17}, speech recognition \cite{DBLP:conf/iscas/SkowronskiH07,DBLP:conf/icnc/ZhaoYCS15} and emotion recognition \cite{DBLP:conf/pit/SchererOSP08,DBLP:conf/annpr/TrentinSS10,DBLP:conf/cisda/SalehMKW15,DBLP:journals/nn/BozhkovKG16}. Meanwhile, there are also many applications based on 3D data, like 3D motion pattern indexing \cite{DBLP:conf/psivt/Bacic15,DBLP:conf/iconip/Bacic16,DBLP:conf/icann/TanisaroLSPH17}, activity /gesture recognition \cite{DBLP:conf/esann/JirakBW15,DBLP:conf/esann/MiciHW16,DBLP:conf/cse/SchaettiSC16,DBLP:conf/icecsys/WangLAH19} and human eye movements \cite{DBLP:conf/ifip12/Koprinkova-Hristova18}.

\begin{itemize}[leftmargin=10 pt]
    \item \textbf{Nature language processing}.
\end{itemize}

In nature language processing (NLP) field, sentence processing \cite{DBLP:conf/icann/Frank06,DBLP:journals/connection/FrankJ10}, existing methods have applied ESN idea in the tasks of grammatical structure learning \cite{DBLP:journals/nn/TongBCC07,DBLP:journals/finr/HinautPPD14,DBLP:conf/ssci/LiuLJ19}, probabilistic language modeling \cite{DBLP:conf/icann/HommaH13}, word embeddings \cite{DBLP:conf/inista/SimovKPO19} and word sense disambiguation \cite{DBLP:conf/icann/PopovKSO19}.

\begin{itemize}[leftmargin=10 pt]
    \item \textbf{Robotics}.
\end{itemize}

In the field of agent (robot) control, robot navigation/localization \cite{DBLP:conf/robio/HartlandB07,DBLP:conf/wirn/ChessaGGM13}, robotics trajectory control \cite{DBLP:conf/cira/SongF09,DBLP:conf/rita/ValenciaVF13,DBLP:conf/cifer/MacielGSB14} and evolutionary Robotics \cite{DBLP:conf/cec/HartlandBS09} have been the applications of ESNs.

\begin{itemize}[leftmargin=10 pt]
    \item \textbf{Ambient-assisted-living applications}.
\end{itemize}

Ambient assisted living (AAL) task aims to recognize the behavior of humans in their every-day environments. For example, by monitoring the regularity of people activities, enhancing the degree of personalization of smart home services. Wireless sensor networks is a mean to gather relevant data for the purpose of modeling the user’s behavior and vast amount of temporal data is generated through the interaction of humans with the sensors. In this context, the interest in the adoption of ESN methodologies to discover relevant patterns from streams of sensorial information is constantly increasing. \cite{DBLP:conf/aiia/GallicchioM17}

\begin{table*}[t]
\caption{Category of methods related to ESN}\label{tb:category}
\centering
\begin{tabular}{|l|l|l|l|l|}
\hline
Category &Subcategory &Researches &Achievements &Feature works\\
\hline

\multirow{6}{*}{Basic ESN}  &\shortstack[l]{Network\\analysis } &\shortstack[l]{\cite{JaegerShortTerm} \cite{DBLP:journals/asc/MaCWY14} \cite{DBLP:conf/icann/BarancokF14} \cite{DBLP:journals/nn/FarkasBG16}\\ \cite{DBLP:journals/nn/KoryakinLB12} \cite{DBLP:conf/ausai/StockdillN16}} &\shortstack[l]{ESP,STM,MC,EOC...\\ \quad} &\shortstack[l]{More evaluation metrics;\\Relations between properties.}\\
\cline{2-5}

&\shortstack[l]{Dynamic\\weights} &\shortstack[l]{\cite{DBLP:conf/bioadit/MayerB04} \cite{DBLP:conf/icann/HajnalL06} \cite{DBLP:conf/ijcnn/FanWJ17} \cite{DBLP:conf/icann/BabinecP07}\\ \cite{DBLP:conf/esann/BoedeckerOMA09} \cite{DBLP:journals/nca/QiaoWY19}} &\shortstack[l]{Self-prediction,critical,\\ reinforcement,IP...} &\shortstack[l]{Auto-ML;\\Simplification complexity.}\\
\cline{2-5}

&\shortstack[l]{Reservoir\\connections} &\shortstack[l]{\cite{DBLP:journals/tnn/RodanT11} \cite{DBLP:journals/jzusc/SunCLCL12} \cite{DBLP:conf/cscwd/SunYZWX18} \cite{DBLP:journals/nn/CuiFCLL14}\\ \cite{DBLP:conf/ideal/Fan-junY16} \cite{DBLP:journals/tnn/QiaoLHL17}} &\shortstack[l]{DLR,DLRB,SCR...\\ \quad} &\shortstack[l]{Dynamic study;\\Relation analysis.}\\
\cline{2-5}

&\shortstack[l]{Multiple\\reservoirs\\ \quad} &\shortstack[l]{\cite{DBLP:conf/icmlc/ChenMP10} \cite{DBLP:journals/tie/HanL13} \cite{DBLP:conf/itsc/SerLBV19} \cite{DBLP:journals/access/HanJLD19}\\ \cite{DBLP:journals/tnn/XiaJHPM11} \cite{DBLP:conf/ijcnn/RachezH12} \cite{DBLP:journals/nn/BoccatoLAZ12} \cite{2011Architectural}\\ \cite{DBLP:journals/access/ZhangZT19} \cite{DBLP:journals/jfi/YaoW19} \cite{DBLP:journals/nn/XueYH07}} &\shortstack[l]{Wide,growing...\\ \quad\\\quad\\\quad} &\shortstack[l]{Auto-ML;\\Dynamic study;\\Relation analysis.}\\
\cline{2-5}

&\shortstack[l]{Hyper-\\parameters\\optimization\\\quad\\\quad} &\shortstack[l]{\cite{DBLP:journals/nn/VenayagamoorthyS09} \cite{DBLP:conf/iscid/CaiFWGZ18} \cite{DBLP:conf/iwcmc/LuoZSS20} \cite{DBLP:journals/nn/VenayagamoorthyS09}\\\cite{DBLP:journals/cin/MartinR15} \cite{DBLP:conf/isda/Basterrech13} \cite{B2003Boosting} \cite{DBLP:conf/ijcnn/MaatGP18}\\ \cite{DBLP:conf/esann/CerinaFS19} \cite{2004Identification} \cite{DBLP:conf/ijcnn/FerreiraL10} \cite{DBLP:conf/iconac/LiuZ18}\\\cite{DBLP:journals/access/BalaIISO20} \cite{DBLP:journals/sensors/HuangLC20} \cite{DBLP:conf/ijcnn/AkiyamaT19} \cite{DBLP:journals/nn/ThiedeP19}\\\cite{DBLP:journals/ijon/ZhangQCL20} \cite{DBLP:journals/ijon/OzturkCI20} \cite{DBLP:journals/ijon/LiuSLYCZ20a}} &\shortstack[l]{GA,PSO,SA,DE...\\ \quad\\\quad\\\quad\\\quad\\\quad} &\shortstack[l]{Simplification complexity;\\Improve universality;\\More practical.\\\quad\\\quad}\\
\cline{2-5}

&\shortstack[l]{Training\\phase\\designs\\\quad\\\quad} &\shortstack[l]{\cite{DBLP:conf/gecco/JiangBS08} \cite{DBLP:conf/ae/DevertBS07} \cite{DBLP:journals/neco/ShutinZKP12} \cite{DBLP:conf/bica/GoudarziS14}\\ \cite{DBLP:journals/nn/ScardapaneWP16} \cite{DBLP:journals/ar/ChewKN15} \cite{DBLP:journals/nn/ScardapaneWP16} \cite{DBLP:conf/ssp/CouilletWSA16}\\ \cite{DBLP:journals/cogcom/LokseBJ17} \cite{DBLP:journals/access/QiaoWYG18} \cite{DBLP:journals/access/YangZAWQ18} \cite{DBLP:conf/icann/NguyenKJK18}\\ \cite{DBLP:journals/tcyb/XuHQL19} \cite{DBLP:journals/eaai/WangZWW19} \cite{DBLP:journals/nn/YangQANW19} \cite{DBLP:conf/icann/LukoseviciusU19}\\ \cite{DBLP:journals/eaai/LuXC20} \cite{DBLP:journals/tcas/LuoFSAY20} \cite{DBLP:journals/tie/LiZO20} \cite{DBLP:conf/ijcnn/LiT20}} &\shortstack[l]{Ill-posed,over-fitting\\\quad\\\quad\\\quad\\\quad\\\quad\\\quad
} &\shortstack[l]{Simplification complexity;\\Improve universality;\\More practical.\\\quad\\\quad\\\quad}\\
\hline

\multirow{2}{*}{Deep ESN} &\shortstack[l]{Network\\analysis} &\shortstack[l]{\cite{DBLP:conf/esann/GallicchioM16} \cite{Gallicchio2017Deep} \cite{DBLP:conf/ijcnn/GallicchioM18} \cite{DBLP:journals/corr/abs-1712-04323}\\ \cite{DBLP:conf/esann/GallicchioM16} \cite{DBLP:journals/corr/GallicchioMP17} \cite{DBLP:journals/cogcom/GallicchioM17}} & \shortstack[l]{ESP,STM,MC,EOC...\\ \quad} & \shortstack[l]{More evaluation metrics;\\Relations between properties.}\\
\cline{2-5}

&\shortstack[l]{Structure\\designs\\ \quad} &\shortstack[l]{\cite{DBLP:conf/esann/BianchiSLJ18} \cite{DBLP:journals/nn/GallicchioMP18} \cite{DBLP:journals/corr/abs-1711-05255} \cite{DBLP:journals/isci/MaSC20}\\\cite{DBLP:journals/tfs/ZhangSWL0020} \cite{DBLP:journals/eaai/ArenaPS20} \cite{DBLP:journals/asc/BoWZ20} \cite{DBLP:journals/corr/abs-1908-08380}\\\cite{DBLP:journals/corr/abs-1808-00523} \cite{DBLP:journals/nca/KimK20} } &\shortstack[l]{Stacked,wide,cross...\\ \quad\\ \quad} & \shortstack[l]{Auto-ML;\\Dynamic study;\\Relation analysis.}\\
\hline

\multirow{2}{*}{Combinations}  &\shortstack[l]{Non-DL\\ \quad}&\shortstack[l]{\cite{DBLP:journals/tnn/ShiH07} \cite{DBLP:journals/cogcom/ScardapaneU17} \cite{DBLP:journals/nca/PengLLP14} \cite{DBLP:journals/ijon/GallicchioM13} \\\cite{2018Deep} \cite{DBLP:conf/ijcnn/GallicchioM10} \cite{DBLP:conf/aaai/GallicchioM20}} &\shortstack[l]{SVM,ARIMA,\\Tree,Graph} &\shortstack[l]{More ML methods;\\Unstructured data}\\
\cline{2-5}

&\shortstack[l]{DL\\ \quad \\ \quad\\ \quad\\ \quad\\ \quad\\ \quad\\ \quad\\ \quad\\ \quad\\ \quad\\ \quad\\ \quad} &\shortstack[l]{\cite{DBLP:conf/icann/BabinecP06} \cite{DBLP:conf/ciasg/JayawardeneV14} \cite{DBLP:conf/esann/RodanT11} \cite{DBLP:conf/isnn/WangPP11}\\ \cite{DBLP:journals/isci/SunJYTLX19} \cite{DBLP:journals/kbs/WangWXWZ20} \cite{DBLP:journals/ijon/ChouikhiAHA19} \cite{DBLP:conf/icann/AkiyamaT19}\\ \cite{DBLP:conf/icann/PopovKSO19} \cite{DBLP:conf/ciarp/LopezVAG17} \cite{DBLP:journals/tim/YangZL20} \cite{DBLP:conf/ijcnn/WangJH20}\\ \cite{DBLP:conf/inista/SarliGM20} \cite{DBLP:journals/access/ZhengQLXZM20} \cite{DBLP:conf/ijcnn/LiT20a} \cite{DBLP:journals/access/ZhangZL19}\\ \cite{DBLP:conf/robio/ZhangCLZ18} \cite{DBLP:journals/tr/FinkZW15} \cite{DBLP:journals/iotj/SunMLWLWG20} \cite{DBLP:journals/ria/ChengZ19}\\ \cite{DBLP:journals/kbs/SunLLHL17} \cite{DBLP:conf/icann/SzitaGL06} \cite{DBLP:journals/corr/abs-2010-05449} \cite{DBLP:conf/esann/Oubbati11}\\ \cite{DBLP:conf/iros/SchmidtBP14} \cite{DBLP:conf/rita/MatsukiS17} \cite{DBLP:conf/smc/Koprinkova-HristovaOP10} \cite{DBLP:conf/esann/Oubbati11}\\ \cite{DBLP:conf/iconip/SiqueiraBAL12}  \cite{DBLP:conf/ijcnn/CarvalhoSFB18} \cite{DBLP:journals/eaai/RibeiroRS20}} &\shortstack[l]{MLP,AE,GAN,RNN,\\CNN,DBN,RL,ELM...\\ \quad\\ \quad\\ \quad\\ \quad\\ \quad\\ \quad\\ \quad\\ \quad} &\shortstack[l]{More ML methods;\\Unstructured data;\\Simplification complexity;\\Improve accuracy;\\More practical.\\ \quad\\ \quad\\ \quad}\\
\hline

\end{tabular}
\end{table*}

\begin{table*}[!ht]
\caption{Applications of methods related to ESN}\label{tb:applications}
\centering
\begin{tabular}{|l|l|l|l|l|}
\hline
Category &Applications &Researches &Achievements &Feature works\\
\hline

\multirow{5}{*}{\shortstack[l]{Abnormal\\data}} &\shortstack[l]{Noises\\ \quad\\ \quad\\ \quad} &\shortstack[l]{\cite{Xue2007Decoupled}\cite{2007Online} \cite{DBLP:journals/ijon/ShengZLW12} \cite{DBLP:journals/isci/XuHL18}\\ \cite{DBLP:conf/annpr/SennK20} \cite{DBLP:journals/ijon/RigamontiBZRGP18} \cite{DBLP:conf/ifsa/LiuZW13} \cite{DBLP:journals/sj/YangZ20}\\ \cite{DBLP:journals/tnn/LiHW12} \cite{DBLP:journals/asc/ShenCZYJ18} \cite{DBLP:conf/ijcnn/ZhangGWC18} \cite{DBLP:journals/ijon/GuoWCX17}} & \shortstack[l]{Noises\\Outliers\\ \quad} & \shortstack[l]{Wave oriented;\\Data enhancement.\\ \quad}\\
\cline{2-5}

&\shortstack[l]{Imbalance\\ \quad} &\shortstack[l]{\cite{DBLP:journals/nca/ChenZHHS20} \cite{DBLP:journals/access/WangWLZW19} \cite{DBLP:journals/tnn/HanX18} \cite{DBLP:journals/nca/QiaoWY19}\\ \quad} &\shortstack[l]{Imbalanced\\Missing} & \shortstack[l]{Unequal intervals;\\Other data structure.}\\
\cline{2-5}

&\shortstack[l]{Dynamic \\data format} &\shortstack[l]{\cite{DBLP:conf/icmla/TanisaroH16} \cite{DBLP:journals/ijista/Song-linX17} \cite{DBLP:journals/nca/HuangQZWS18} \cite{DBLP:journals/ajiips/TanakaNH19}\\ \cite{DBLP:journals/nn/YaoWZ19} \cite{DBLP:journals/access/XiangLZPWL16} \cite{DBLP:conf/clef/Schaetti18b}} &\shortstack[l]{Temporal\\Two direction} & \shortstack[l]{Real-time\\Changing}\\
\cline{2-5}

&\shortstack[l]{Expert} &\shortstack[l]{\cite{DBLP:conf/iccS/DoanPM19} \cite{DBLP:journals/corr/abs-2011-06769}} &\shortstack[l]{Physic law} & \shortstack[l]{Other knowledge}\\
\cline{2-5}

&\shortstack[l]{High\\dimensional} &\shortstack[l]{\cite{DBLP:conf/cse/SchaettiSC16} \cite{DBLP:conf/esann/MiciHW16} \cite{DBLP:conf/acii/WenLL15} \cite{DBLP:journals/tnn/DuanW16}\\ \cite{DBLP:journals/tnn/XiaJM15}} &\shortstack[l]{2D,3D,4D\\ \quad} & \shortstack[l]{Unstructured data.\\\quad}\\
\hline

\multirow{8}{*}{\shortstack[l]{Real-world\\tasks}}  &\shortstack[l]{Industrial\\ \quad \\ \quad\\ \quad\\ \quad\\ \quad\\ \quad\\ \quad\\ \quad\\ \quad \\ \quad\\ \quad\\ \quad\\ \quad\\ \quad} &\shortstack[l]{\cite{DBLP:conf/ifip12/DettoriMCS20} \cite {DBLP:conf/eann/CollaMDCM19} \cite{DBLP:conf/ijcnn/MansoorGM20} \cite{DBLP:conf/ifip12/DettoriMCS20}\\ \cite{DBLP:journals/mcs/MorandoJHGZ17} \cite{DBLP:journals/tie/LiZO20} \cite{DBLP:conf/iecon/MorandoJGZH13} \cite{energy}\\ \cite{DBLP:journals/sensors/SanchezAOC18} \cite{DBLP:journals/eaai/JordanouAC19} \cite{DBLP:journals/tim/YangZL20} \cite{DBLP:conf/ijcnn/DaiVH09}\\ \cite{DBLP:conf/cies/AquinoNSLCLF14} \cite{DBLP:conf/ideal/Dorado-MorenoGS18} \cite{DBLP:conf/ciasg/JayawardeneV14} \cite{DBLP:conf/iciai/ZhangLCYZL19}\\ \cite{DBLP:conf/iciai/ZhangLCYZL19} \cite{DBLP:journals/nn/YaoWZ19} \cite{DBLP:conf/icra/SalmenP05} \cite{DBLP:conf/ipsn/ObstWP08}\\ \cite{DBLP:conf/dcoss/ChangTB09} \cite{DBLP:conf/ijcnn/WoottonDH18} \cite{DBLP:journals/tii/LongZL20}\cite{DBLP:journals/tfs/ZhangSWL0020}\\ \cite{DBLP:journals/eaai/XuBAZ20} \cite{DBLP:journals/eaai/LuXC20} \cite{DBLP:journals/access/BalaIISO20} \cite{DBLP:journals/nn/Venayagamoorthy07}\\ \cite{DBLP:journals/nn/AntoneloCF17} \cite{DBLP:conf/wises/MathewsP08} \cite{DBLP:conf/isnn/QinHZ09} \cite{DBLP:books/sp/09/KrauseBDS09}\\ \cite{DBLP:conf/isda/SongFKL09} \cite{DBLP:conf/rose/HarischandraD12} \cite{DBLP:conf/ijcnn/KuwabaraNKBGCP12} \cite{DBLP:journals/nca/PengLLP14}\\ \cite{DBLP:journals/nn/BianchiSURS15} \cite{DBLP:journals/iet-com/GideonNT17} \cite{DBLP:journals/esticas/BaiYZJL20}} &\shortstack[l]{Energy\\Manufacturing\\ \quad\\ \quad\\ \quad\\ \quad\\ \quad\\ \quad\\ \quad\\ \quad\\ \quad\\ \quad\\ \quad} &\shortstack[l]{Others \\ \quad\\ \quad\\ \quad\\ \quad\\ \quad\\ \quad\\ \quad\\ \quad\\ \quad\\ \quad\\ \quad\\ \quad\\ \quad}\\
\cline{2-5}

&\shortstack[l]{Medical\\ \quad \\ \quad\\ \quad\\ \quad\\ \quad\\ \quad\\ \quad\\ \quad} &\shortstack[l]{\cite{DBLP:journals/isci/SunJYTLX19} \cite{DBLP:journals/cbm/KimJ19} \cite{DBLP:conf/ijcnn/FouratiAJA20} \cite{DBLP:conf/embc/AyyagariJW14} \\\cite{DBLP:conf/embc/AyyagariJW15} \cite{DBLP:conf/iconip/FouratiAASA17} \cite{DBLP:journals/nca/RenDW19} \cite{DBLP:journals/nca/RenDW20} \\\cite{DBLP:conf/icann/RuizGJM20} \cite{DBLP:conf/idaacs/PetrenasML11} \cite{DBLP:journals/tbe/PetrenasMSL12} \cite{DBLP:conf/amia/FongMRR14} \\\cite{DBLP:conf/mibam/WismullerDAWHN15} \cite{DBLP:conf/esann/GallicchioMP18} \\\cite{DBLP:journals/artmed/LacySL18} \cite{DBLP:journals/midm/VerplanckeLSBTMD10} \cite{DBLP:conf/cicare/WajidHL14} \cite{DBLP:conf/cinc/AlfarasVG19} \\\cite{DBLP:journals/jms/Al-MaaitahA18} \cite{DBLP:journals/nca/KainzBAA17} \cite{DBLP:journals/ijon/LiTWW20}} &\shortstack[l]{EEG,ECG,MRI\\Cancer\\Sepsis\\ \quad\\ \quad\\ \quad} &\shortstack[l]{Others\\ \quad\\ \quad\\ \quad\\ \quad\\ \quad\\ \quad}\\
\cline{2-5}

&\shortstack[l]{Financial\\ \quad } &\shortstack[l]{\cite{DBLP:journals/nca/KimK20} \cite{DBLP:conf/pakdd/LinYS08} \cite{DBLP:journals/eswa/LinYS09} \cite{DBLP:journals/eswa/LinYS11}\\ \cite{DBLP:conf/iconip/LiuSLFCZD18}} &\shortstack[l]{Stock\\Prize \quad} &\shortstack[l]{ Others\\ \quad}\\
\cline{2-5}

&\shortstack[l]{AAL } &\shortstack[l]{\cite{DBLP:conf/aiia/GallicchioM17}} &\shortstack[l]{Representation \\ \quad} &\shortstack[l]{Others}\\
\cline{2-5}

&\shortstack[l]{Spatio-\\temporal\\ \quad\\ \quad\\ \quad\\ \quad\\ \quad\\ \quad} 
&\shortstack[l]{\cite{DBLP:journals/tnn/SohD15} \cite{DBLP:conf/icra/SchaffernichtBL17} \cite{DBLP:journals/corr/abs-1806-10728} \cite{DBLP:journals/corr/abs-2004-08170}\\ \cite{DBLP:conf/icnc/AnSZ11} \cite{DBLP:conf/itsc/SerLBV19} \cite{DBLP:journals/corr/abs-2004-08170} \cite{DBLP:journals/ijon/SongWS20}\\ \cite{DBLP:journals/nca/KimK20} \cite{DBLP:journals/ejes/MathewsP09} \cite{DBLP:conf/esann/SchwenkerL09} \cite{DBLP:conf/isda/BasterrechZPM13}\\ \cite{DBLP:journals/asc/LiWLFL20} \cite{DBLP:conf/ecc/BasterrechB14} \cite{DBLP:conf/eann/KmetK15} \cite{DBLP:journals/tii/WuLX21}\\ \cite{DBLP:journals/nca/KimK20} \cite{DBLP:journals/tnn/XuYHQL19} \cite{DBLP:journals/nca/HuangQZWS18} \cite{DBLP:conf/ijcnn/SacchiOPCS07}\\ \cite{DBLP:conf/ideal/SiqueiraBAF12} \cite{DBLP:conf/iconip/SiqueiraBAL12} \cite{DBLP:conf/ijcnn/Qiao09} \cite{DBLP:conf/ciarp/LopezVAG17}\\ \cite{DBLP:conf/amcc/ChitsazanFNT17}} &\shortstack[l]{Traffic\\Astronomy\\Meteorology\\ \quad\\ \quad\\ \quad\\ \quad} &\shortstack[l]{Others\\\quad\\ \quad\\ \quad\\ \quad\\ \quad\\ \quad\\ \quad}\\
\cline{2-5}

&\shortstack[l]{CV\\ \quad\\ \quad\\ \quad\\ \quad\\ \quad\\ \quad\\ \quad \\ \quad\\ \quad} 
&\shortstack[l]{\cite{DBLP:conf/ijcnn/SouahliaBBC16} \cite{DBLP:journals/cogcom/MeftahLB16} \cite{DBLP:conf/icaci/SouahliaBBC17} \cite{DBLP:conf/mesas/DonkorSB18} \\\cite{DBLP:journals/concurrency/SouahliaBBABC20} \cite{DBLP:journals/tnn/DuanW16} \cite{DBLP:conf/acii/WenLL15} \cite{DBLP:conf/isnn/SquartiniCRP07}\\ \cite{DBLP:journals/nn/GallicchioMP18} \cite{DBLP:conf/esann/GallicchioMP19} \cite{DBLP:conf/ccis/SunCLCL12} \cite{DBLP:conf/icmla/RoychowdhuryM18} \\\cite{DBLP:journals/cogcom/ScardapaneU17} \cite{DBLP:conf/iscas/SkowronskiH07} \cite{DBLP:conf/icnc/ZhaoYCS15} \cite{DBLP:conf/pit/SchererOSP08} \\\cite{DBLP:conf/annpr/TrentinSS10} \cite{DBLP:conf/cisda/SalehMKW15}\cite{DBLP:journals/nn/BozhkovKG16} \cite{DBLP:conf/psivt/Bacic15}\\ \cite{DBLP:conf/iconip/Bacic16} \cite{DBLP:conf/icann/TanisaroLSPH17} \cite{DBLP:conf/esann/JirakBW15} \cite{DBLP:conf/esann/MiciHW16}\\ \cite{DBLP:conf/cse/SchaettiSC16} \cite{DBLP:conf/icecsys/WangLAH19} \cite{DBLP:conf/ifip12/Koprinkova-Hristova18}} &\shortstack[l]{Image segmentation\\Emotion recognition\\gesture recognition\quad\\ \quad\\ \quad\\ \quad\\ \quad\\ \quad\\ \quad} &\shortstack[l]{Others\\ \quad\\ \quad\\ \quad\\ \quad\\ \quad\\ \quad\\ \quad\\ \quad\\ \quad}\\
\cline{2-5}

&\shortstack[l]{NLP\\ \quad } &\shortstack[l]{\cite{DBLP:conf/icann/Frank06} \cite{DBLP:journals/connection/FrankJ10} \cite{DBLP:journals/nn/TongBCC07} \cite{DBLP:journals/finr/HinautPPD14}\\ \cite{DBLP:conf/ssci/LiuLJ19} \cite{DBLP:conf/icann/HommaH13} \cite{DBLP:conf/inista/SimovKPO19} \cite{DBLP:conf/icann/PopovKSO19}} &\shortstack[l]{Sentence processing\\Embedding} &\shortstack[l]{Others\\ \quad}\\
\cline{2-5}

&\shortstack[l]{Robotics\\ \quad } &\shortstack[l]{\cite{DBLP:conf/robio/HartlandB07} \cite{DBLP:conf/wirn/ChessaGGM13} \cite{DBLP:conf/cira/SongF09} \cite{DBLP:conf/rita/ValenciaVF13} \\\cite{DBLP:conf/cifer/MacielGSB14} \cite{DBLP:conf/cec/HartlandBS09}} &\shortstack[l]{Navigation\\Localization} &\shortstack[l]{Others\\ \quad}\\
\hline
\end{tabular}
\end{table*}

\section{Discussions} \label{sec:Discussions}

\subsection{Challenges and open questions} \label{sec:Challenges}
According to the analysis of the above research progress, there are many breakthroughs and good applications about ESN since it was proposed. However, some open questions still need to be solved.

\begin{enumerate}[leftmargin=10 pt]
    \item \textbf{How the parameters of reservoir network structure effect the tasks? individually and collectively? generally and specifically?}
    
    The designs of network structures related to many considerations, such as hyper-parameters initialization and network connection. Some works have studied that the influence of three basic hyper-parameters and network topology to the final tasks. And they tried to suggest selections. But most of them tested the impact by changing one variable and fixing other variables. Meanwhile, in different application scenarios or downstream tasks, the impact seems to be different. Thus, three open questions are: 1) Do the parameters related to network design interact with each other? 2) What are their common and separate relations to the final tasks? 3) How the design affects common tasks and specific applications? 
    
    \item \textbf{What is the best setting of reservoir network structure? Is there an optimal reservoir network structure that we can construct from each task? How to connect different dynamics and different best reservoir structure?}
    
    Although some achievements have been made in the selection of hyper-parameters, but the best setting of weight matrix initialization and network topology is still an open question. Some theories have proved the setting method on some chaotic systems, like Mackey-Glass system, but for specific tasks, such as industrial and medical applications, the network settings are often determined by experience and experiments with not much theoretical guidance.
    
    Meanwhile, before we considered how to design a best network for a task, we can't prove whether the optimal design exists. Even if it does, it is unknown whether the idea of ESN can achieve it. To go further, we are not sure about the definition of the "best", both the results of the task and the performance of the network need to be considered. For example, memory capacity and active information storage are both the measure to assess the computational properties of ESNs. Under different assessments, different optimal structures may be obtained.
    
    \item \textbf{How to ensure the accuracy of ESN in fast calculation? On the contrary? Can ESNs be the plain substitute for BP RNN in the future?}
    
    At present, there is little or no literature demonstrate that ESNs can completely replace other RNN methods. In most of complex tasks such as speech recognition, deep learning methods with gradient descent by BP algorithm still have the best accuracy due to their highly nonlinear of ‘black-box’. However, ESNs could be an alternative in some tasks that are designed to run on smaller, constrained devices where the size or performance of RNN is limited. But the complexity of ESNs is not always small. In order to get the best structure, some complex optimization algorithms are used, leading to heavy computational cost, which will weaken the competitiveness like training time of ESNs. However, reducing the complexity of the optimization algorithm often goes against improving the accuracy of the results. How to design a fast and accurate optimization algorithm is a problem. 
\end{enumerate}

\subsection{Opportunities and future works} \label{sec:Opportunities}

Based on the above three open questions, we propose some possible directions for future work. 

\begin{enumerate}[leftmargin=10 pt]
    \item \textbf{Study the relations between the parts of ESN.}
    
    The parts of ESN include inputs like data type and frequency, reservoir structure like network connections and number of layers, initialization parameters like matrix sparsity and scale, and outputs like results accuracy. Learning their relations is helpful to better design the network structure. For example, determination of the relations between input datasets and hyper-parameters helps adaptively adjust the structure and weight of reservoir; Studies on the frequency analysis of DeepESN dynamics allowed to develop an algorithm for the automatic setup of (the number of layers of) a DeepESN; The analysis of dynamic relationship and representation within ESN by using different technologies, like dimensionality reduction methods, time-varying graphs and other recurrence analysis tools, also provides insights for understanding the input system possibly.
    
    \item \textbf{Study the optimization algorithm which is not only accurate but also fast.}
    
    Some optimization algorithms discard the advantage of fast computation of ESN when they pursue high accuracy. How to reduce the complexity of the optimization algorithm is a necessary future work. Meanwhile, the studies about speeding up the training process of ESNs, such as designing mini-batches, parallelizing calculation and taking advantage of GPU-based devices possibility, are required. Meanwhile, some basic issues still need to be addressed, like over-fitting problem.
    
    \item \textbf{Study the method of automatically generating the optimal network structure of ESNs.}
    
    The method can not only improve the traditional parameter optimization method, but also refer to the deep learning method. For example, the techniques that can automate machine learning models both structurally and parametrically under the AutoML paradigm. Possible work can import the idea in AutoML area to reservoir-based realm and leverage the advances of modular topology. A fully modular ensemble architecture of ESN is also worth studying. 
    
    \item \textbf{Study the evaluation metrics of network property.}
     
    The network structure with the highest prediction accuracy does not necessarily have the best network performance. And different assessments lead to different optimal structure. The relations between network performance described by existing assessments needs to be studied. For example, a possible further research is the relationship between the maximal MC, the loss of the echo state property and the edge of chaos, which is rather complex nowadays. Another work worth studying is the relevance of network property and task outcome and the network dynamics in different tasks. These two kinds of studies can not only help to complete the task better, but also provide insights for network dynamic understanding and data relationship.
    
    \item \textbf{From theory to practice.}
    
    At present, many existing practical aspects for successfully applying ESNs are not universal and should be filtered depending on a particular task. For example, most of the optimization algorithms are experimented on the simulation system, like MG system. But in the real-world industrial practice, the stability and regularity of the system cannot be guaranteed. Some rules and expert knowledge need to be considered. Although some methods are proposed for practical tasks, they are also not the only possible approaches, and can most likely be improved upon. Thus, the future work not only needs to put forward the solid theory, but also needs to put the method into practice.
    
    \item \textbf{From specific applications to universal applications.}
    
    Some work is proposed for a particular point, but it is not suitable for most scenarios. For example, a noise-robust ESN can process univariable time series, but it can't deal with multivariable data. However, in the real world, multivariate time series are common. We do not deny the design of solutions to specific difficulties, we think that the scope of application of the method is a worthy research direction to improve the algorithm.
    
    \item \textbf{From traditional applicable data to diversified data}
    
    Since ESNs was proposed, it has been used to model time series data. Although some methods propose to model noisy sequence data, there are many other irregularities in time series data such as unequal intervals between observations in two adjacent time points and different sampling rates between different time series. Meanwhile, the basic ESNs model short-term memory with no explicit mechanism to connect the multi-span memory or learn longer memory. But time series may have different information in different spans, and have strong correlation with long-term data, such as periodic data. Therefore, future developments can be boosted by an enriched representation of the input dynamics, exploiting the time-scale differentiation in time granularity. At present, there are methods to apply ESN to tree-structured data and graph-structured data. In the future, ESN may also be used in unstructured data.
    
    \item \textbf{Develop DeepESNs.}
    
    DeepESNs combines the advantages of both ESNs and deep learning. The former pursues conciseness and effectiveness, but the latter focuses on the capacity of learning complex features. Thus, there is a gap between these two approaches, and a potentially future direction is to balance the efficiency and the feature learning capacity. But existing works focus more on the designs of stacked layers but ignore the basic theory. But better understanding the constitutive meaning of stacking layers of DeepESNs is the foundations for further model variants and extensive applications. Thus, more works about the basic theory are required.
    
\end{enumerate}

\section{Conclusion} \label{sec:Conclusion}

In this paper, we categorize the ESN-based methods to basic ESNs, DeepESNs and combinations, then analyze them from the perspective of theoretical studies, network designs and specific applications. Finally, we discuss the challenges and opportunities this area, summarize three open questions and propose eight possible future works. This review aims to summarize the current ESN-based works and guide the future works with potential research directions.

\bibliographystyle{unsrt} 
\bibliography{ESNs}

\end{document}